%% file: main.tex
\documentclass{article}

\input{preamble}

\input{math_commands.tex}

\usepackage[accepted]{icml2026}

\icmltitlerunning{PolySAE: Modeling Feature Interactions in Sparse Autoencoders via Polynomial Decoding}

\begin{document}

\twocolumn[
  \icmltitle{PolySAE: Modeling Feature Interactions in Sparse Autoencoders\\ via Polynomial Decoding}




  \begin{icmlauthorlist}
    \icmlauthor{Panagiotis Koromilas}{cyi,uoa}
    \icmlauthor{Andreas D. Demou}{cyi}
    \icmlauthor{James Oldfield }{ox}
    \icmlauthor{Yannis Panagakis}{uoa,arch}
    \icmlauthor{Mihalis A. Nicolaou}{ucy,cyi}
  \end{icmlauthorlist}

  \icmlaffiliation{cyi}{The Cyprus Institute}
  \icmlaffiliation{uoa}{University of Athens}
  \icmlaffiliation{ox}{University of Oxford}
  \icmlaffiliation{arch}{Archimedes AI/Athena Research Center}
  \icmlaffiliation{ucy}{University of Cyprus}

  \icmlcorrespondingauthor{Panagiotis Koromilas}{pakoromilas@di.uoa.gr}
  
  \icmlkeywords{Sparse Autoencoders, Mechanistic Interpretability, Feature Interactions, Polynomial Networks, Representation Learning}

  \vskip 0.3in
]



\printAffiliationsAndNotice{}  

\input{sections/0.abstract}
\input{figures/motivation}
\begin{figure*}
    \centering
    \includegraphics[width=1.0\linewidth]{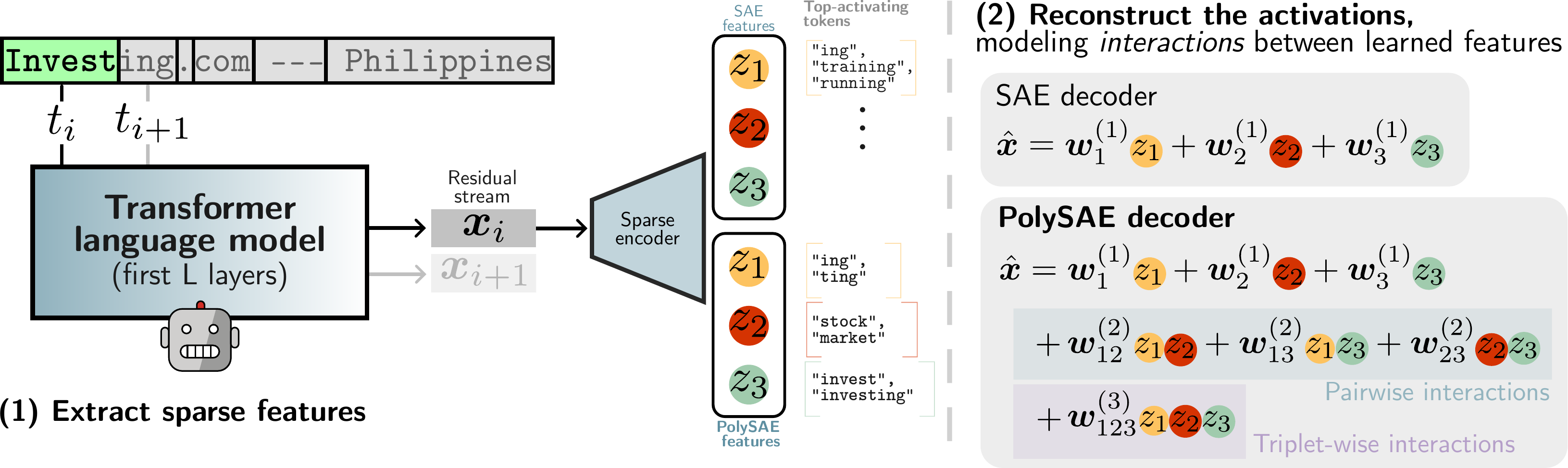} 
    \caption{\textbf{An overview of PolySAE}: (1) sparse latent features are first extracted with a standard SAE encoder. (2) Activations in the residual stream are then reconstructed by modeling 2nd- and 3rd-order interactions in addition to the standard linear component. The example ``\textbf{Invest}ing.com --- Philippines stocks were higher after'' comes from \Cref{tab:polysae_third_order}.}
    \label{fig:placeholder}
\end{figure*}
\input{sections/1.intro}
\input{sections/2.related_work}
\input{sections/4.method}

\input{sections/5.experiments}
\input{sections/6.conclusion}

\section*{Acknowledgements}
This work is supported by the TensorICE project (EXCELLENCE/0524/0407), implemented under the social cohesion programme “THALIA 2021-2027”, co-funded by the European Union through the Research and Innovation Foundation. Yannis Panagakis was supported by the project MIS 5154714 of the National Recovery and Resilience Plan Greece 2.0 funded by the European Union under the NextGenerationEU Program. The authors gratefully acknowledge the EuroHPC Joint Undertaking for awarding this project access to the MareNostrum5 supercomputer, hosted by the Barcelona Supercomputing Center, Spain, under Development Access project ID EHPC-DEV-2025D06-069. This work was also supported by computing time awarded on the Cyclone supercomputer of the High Performance Computing Facility of The Cyprus Institute. 
\section*{Impact Statement}
This work contributes to the field of interpretable machine learning by introducing PolySAE, a method for modeling non-additive feature interactions in sparse autoencoders. By enabling explicit representation of compositional structure while preserving linear, human-interpretable features, this approach advances tools for mechanistic analysis of large language models. Improved interpretability has the potential to support downstream efforts in model auditing, debugging, and safety research by making it easier to identify, analyze, and intervene on meaningful internal representations.

The primary anticipated benefits of this work are methodological and scientific. PolySAE is intended as an analysis tool rather than a deployment-facing component, and it does not directly increase the capabilities of language models. As with other interpretability methods, there is a possibility that insights into internal representations could be misused to more effectively manipulate model behavior, but we do not identify novel or unique risks introduced by this work beyond those already present in the interpretability literature.

Overall, we believe this work has a net positive societal impact by strengthening the technical foundations of interpretability and contributing to the long-term goal of building more transparent, controllable, and trustworthy machine learning systems.

\bibliography{main.bib}
\bibliographystyle{icml2026}

\newpage
\appendix
\onecolumn

\input{sections/appendix}

\end{document}

%% file: preamble.tex
\usepackage[T1]{fontenc}
\usepackage[utf8]{inputenc}

\usepackage{microtype}
\usepackage{xspace}
\usepackage{url}
\usepackage{xcolor}
\usepackage[table]{xcolor}

\usepackage{graphicx}
\usepackage{wrapfig}

\usepackage{subcaption}

\usepackage{booktabs}
\usepackage{array}
\usepackage{tabularx}
\usepackage{multirow}
\usepackage{makecell}
\usepackage{ragged2e}
\usepackage{colortbl}
\usepackage{makecell}

\newcolumntype{L}[1]{>{\RaggedRight\arraybackslash}p{#1}}
\newcolumntype{Y}{>{\RaggedRight\arraybackslash}X}

\usepackage{pgfplots}
\usepgfplotslibrary{groupplots}
\usepgfplotslibrary{colorbrewer}
\usepgfplotslibrary{fillbetween}
\pgfplotsset{compat=1.18}
\usepgfplotslibrary{groupplots,colormaps}
\usepackage{tikz}
\usetikzlibrary{arrows.meta, calc}
\usepackage{pgfplots}
\pgfplotsset{compat=1.18}
\usetikzlibrary{pgfplots.colormaps}
\usepackage{amsmath}
\usepackage{amssymb}
\usepackage{amsfonts}
\usepackage{mathtools}
\usepackage{bm}
\usepackage{amsthm}

\usepackage{enumitem}
\usepackage{tcolorbox}
\usepackage{soul}
\usepackage{todonotes}

\usepackage{hyperref}
\usepackage[capitalize,noabbrev]{cleveref}

\theoremstyle{plain}

\theoremstyle{definition}

\theoremstyle{remark}

\definecolor{ForestGreen}{RGB}{34,139,34}
\definecolor{NavyBlue}{RGB}{0,32,96}
\definecolor{DarkOrange}{RGB}{255,140,0}
\definecolor{decodergreen}{rgb}{0.702, 1.0, 0.502}
\definecolor{hiddenorange}{rgb}{1.0, 0.702, 0.502}
\definecolor{outputgray}{rgb}{0.851, 0.851, 0.851}
\definecolor{AsterixCol}{RGB}{128,128,128}

\newcommand{\eg}{\emph{e.g.}\xspace}
\newcommand{\ie}{\emph{i.e.}\xspace}

\newcommand{\TODO}[1]



%% file: math_commands.tex
\usepackage{amsmath,amsfonts,bm}

\def\eqref#1{equation~\ref{#1}}

\def\1{\bm{1}}

\def\vb{{\bm{b}}}

\def\vd{{\bm{d}}}
\def\ve{{\bm{e}}}

\def\vh{{\bm{h}}}

\def\vm{{\bm{m}}}

\def\vu{{\bm{u}}}

\def\vx{{\bm{x}}}
\def\vy{{\bm{y}}}
\def\vz{{\bm{z}}}

\def\mA{{\bm{A}}}
\def\mB{{\bm{B}}}
\def\mC{{\bm{C}}}
\def\mD{{\bm{D}}}
\def\mE{{\bm{E}}}

\def\mI{{\bm{I}}}

\def\mM{{\bm{M}}}

\def\mQ{{\bm{Q}}}
\def\mR{{\bm{R}}}
\def\mS{{\bm{S}}}

\def\mU{{\bm{U}}}

\def\mGamma{{\bm{\Gamma}}}

\DeclareMathAlphabet{\mathsfit}{\encodingdefault}{\sfdefault}{m}{sl}
\SetMathAlphabet{\mathsfit}{bold}{\encodingdefault}{\sfdefault}{bx}{n}

\def\sR{{\mathbb{R}}}

\newcommand{\cS}{\mathcal{S}}

\newcommand{\R}{\mathbb{R}}



%% file: sections/0.abstract.tex
\begin{abstract}
Sparse autoencoders (SAEs) interpret neural network representations by decomposing activations into sparse combinations of dictionary atoms. However, SAEs assume features combine additively through linear reconstruction, an assumption that cannot capture compositional structure: linear models cannot distinguish whether ``Starbucks'' arises from the composition of ``star'' and ``coffee'' features or merely their co-occurrence. This forces SAEs to allocate monolithic features for compound concepts rather than decomposing them into interpretable constituents. We introduce PolySAE, which extends the SAE decoder with higher-order terms to model feature interactions while preserving the linear encoder essential for interpretability. Through low-rank tensor factorization on a shared projection subspace, PolySAE captures pairwise and triple feature interactions with small parameter overhead (3\% on GPT2). Across four language models and three SAE variants, PolySAE achieves an average improvement of $\sim$8\% in probing F1 while maintaining comparable reconstruction error, and produces 2--10$\times$ larger Wasserstein distances between class-conditional feature distributions. Critically, learned interaction weights exhibit negligible correlation with co-occurrence frequency ($r = 0.06$ vs.\ $r = 0.82$ for SAE feature covariance), suggesting that polynomial terms capture compositional structure largely independent of surface statistics. Finally, the learned interaction directions causally steer model outputs toward the corresponding compositional semantics. Code: \url{https://github.com/pakoromilas/PolySAE}
\end{abstract}

%% file: figures/motivation.tex
%
%
%
\definecolor{axiscolor}{RGB}{130,130,130}
\definecolor{vecblue}{RGB}{50,100,190}
\definecolor{vecred}{RGB}{190,45,45}
\definecolor{vecorange}{RGB}{235,125,10}
\definecolor{planegray}{RGB}{225,232,242}
\definecolor{textgray}{RGB}{75,75,75}
%

\begin{figure}[t]
\centering
\begin{tikzpicture}[>=Stealth, line cap=round, line join=round,
    x={(-0.30cm,-0.18cm)},
    y={(0.58cm,-0.04cm)},
    z={(0cm,0.60cm)},
    scale=1.0]

\begin{scope}[xshift=-2.2cm]
    
    \node[font=\sffamily\bfseries\scriptsize] at (1.4,1.4,4.5) {(a) Additive Interactions};
    
    \fill[planegray, opacity=0.75] (0,0,0) -- (2.8,0,0) -- (2.8,2.8,0) -- (0,2.8,0) -- cycle;
    \draw[axiscolor!50, thin] (0,0,0) -- (2.8,0,0) -- (2.8,2.8,0) -- (0,2.8,0) -- cycle;
    
    \draw[->, axiscolor, line width=0.8pt] (0,0,0) -- (3.3,0,0);
    \draw[->, axiscolor, line width=0.8pt] (0,0,0) -- (0,3.3,0);
    \draw[->, axiscolor, line width=0.8pt] (0,0,0) -- (0,0,2.8);
    
    \node[font=\sffamily\tiny, textgray] at (3.5,-0.5,0) {Famous};
    \node[font=\sffamily\tiny, textgray] at (-0.2,3.5,-0.5) {Beverage};
    \node[font=\sffamily\tiny, textgray, anchor=east] at (-0.15,0,3.0) {Brand};
    
    \draw[dashed, axiscolor!60, thin] (2.1,0,0) -- (2.1,2.1,0);
    \draw[dashed, axiscolor!60, thin] (0,2.1,0) -- (2.1,2.1,0);
    
    \draw[->, vecblue, line width=2pt] (0,0,0) -- (2.1,0,0);
    \node[font=\sffamily\tiny, vecblue] at (2.4,-0.6,0.5) {Star $(z_j)$};
    
    \draw[->, vecblue, line width=2pt] (0,0,0) -- (0,2.1,0);
    \node[font=\sffamily\tiny, vecblue, anchor=south] at (0,2.3,0.15) {Coffee $(z_i)$};
    
    \draw[->, textgray, line width=2.2pt] (0,0,0) -- (2.1,2.1,0);
    \node[font=\sffamily\tiny\bfseries, textgray] at (2.4,2.5,0) {``Star Coffee''};
    
    \node[font=\sffamily\tiny, textgray, align=center, text width=2.3cm] at (1.4,1.4,-1.0) 
        {\textit{Constrained to the}\\\textit{linear span of features}};
    
\end{scope}

\begin{scope}[xshift=2.2cm]
    
    \node[font=\sffamily\bfseries\scriptsize] at (1.4,1.4,4.5) {(b) Multiplicative Interactions};
    
    \fill[planegray, opacity=0.35] (0,0,0) -- (2.8,0,0) -- (2.8,2.8,0) -- (0,2.8,0) -- cycle;
    \draw[axiscolor!40, thin] (0,0,0) -- (2.8,0,0) -- (2.8,2.8,0) -- (0,2.8,0) -- cycle;
    
    \draw[->, axiscolor, line width=0.8pt] (0,0,0) -- (3.3,0,0);
    \draw[->, axiscolor, line width=0.8pt] (0,0,0) -- (0,3.3,0);
    \draw[->, axiscolor, line width=0.8pt] (0,0,0) -- (0,0,2.8);
    
    \node[font=\sffamily\tiny, textgray] at (3.5,-0.5,0) {Famous};
    \node[font=\sffamily\tiny, textgray] at (1.5,3.5,0) {Beverage};
    \node[font=\sffamily\tiny, textgray, anchor=east] at (-0.15,0,3.0) {Brand};
    
    \draw[->, vecblue, line width=1.2pt, opacity=1] (0,0,0) -- (2.1,0,0);
    \draw[->, vecblue, line width=1.2pt, opacity=1] (0,0,0) -- (0,2.1,0);
    
    \coordinate (Base) at (2.1,2.1,0);
    \coordinate (Lifted) at (2.1,2.1,2.1);
    
    \draw[dashed, axiscolor!50, thin] (0,0,0) -- (Base);
    
    
    \fill[axiscolor] (Base) circle (1.5pt);
    
    \draw[->, vecorange, line width=2.5pt] (Base) -- (Lifted);
    \node[font=\sffamily\tiny, vecorange, anchor=west] at (2.3,2.1,1.05) {$z_i \cdot z_j$};
    
    \draw[->, vecred, line width=2.5pt] (0,0,0) -- (Lifted);
    \node[font=\sffamily\scriptsize\bfseries, vecred] at (2,2.1,2.45) {Starbucks};
    
    \node[font=\sffamily\tiny, textgray, align=center, text width=2.3cm] at (1.4,1.4,-1.0) 
        {\textit{Interaction lifts into} \textit{new semantic dimensions}};
    
\end{scope}

\end{tikzpicture}

\caption{\textbf{Semantic Dimension Expansion via Feature Interaction.} 
Consider two semantic directions---\textit{Famous} and \textit{Beverage}---and their associated learned features \textit{Star} and \textit{Coffee}. 
\textbf{(a)}~Additive interactions yield co-occurrence semantics that remain in the original feature span. 
\textbf{(b)}~Multiplicative interactions enable representations to escape this subspace via $z_i \cdot z_j$, lifting into orthogonal dimensions (\textit{Brand}) to capture emergent concepts like \textit{Starbucks}.``Starbucks'' example from \Cref{tab:polysae_second_order}.}
\label{fig:motivation_emergence}
\end{figure}

%% file: sections/1.intro.tex
\section{Introduction}\label{sec:introduction}
As AI systems are increasingly deployed in real-world domains, ensuring their safety and reliability has become a critical challenge \citep{amodei2016concrete,hendrycks2021unsolved,bengio2025international}. Developing interpretable models offers a promising path towards aligning AI with human values: understanding why a model produces a given output enables us to (i) monitor its reasoning \citep{lindsey2025biology}, (ii) debug failure modes \citep{wong2021leveraging}, and (iii) steer away from unwanted behavior \citep{rimsky2024steering}. Mechanistic interpretability pursues this agenda at the level of neural network internals \citep{bereska2024mechanistic}, aiming to uncover interpretable features and circuits within a model and thereby provide principled insights into its behavior.

Sparse Autoencoders (SAEs), grounded in the principles of sparse dictionary learning, have emerged as a leading tool for mechanistic interpretability. SAEs decompose neural network activations to recover human-interpretable features that models typically represent in superposition---encoded in overlapping directions due to limited representational capacity~\citep{elhage2022toy}. This framework has been shown to uncover safety-relevant concepts such as deception, bias, and harmful content, enabling targeted interventions that predictably steer model behavior~\citep{templeton2024scaling}. 

However, recent work has highlighted fundamental limitations of the SAE paradigm due to their reliance on the ``strong'' linear representation hypothesis~\citep{engelsnot,csordas2024recurrent}. Standard SAEs reconstruct activations as weighted sums of independent features, expressing each activation as a linear combination where features contribute additively. This linearity assumption raises a fundamental question: \textit{what level of abstraction do learned features naturally capture?} The answer has direct implications for mechanistic interpretability. If features truly combine linearly, we would expect individual dictionary atoms to represent atomic components, such as morphemes, simple concepts, or basic semantic primitives, that combine through superposition to form complex expressions. Such atomic features would enable transparent circuit analysis and precise interventions on elemental building blocks of meaning.

Yet linguistic theory demonstrates that composition operates non-linearly across multiple levels of language structure. Morphologically, ``administrators'' is not simply the sum of stem and suffix; the combination produces a distinct lexical item with specific syntactic and semantic properties \cite{haspelmath2013understanding}. Semantically, phrasal meanings such as ``kick the bucket' or proper names like ``Starbucks'' (\Cref{fig:motivation_emergence}) \textit{exhibit emergent properties irreducible to their parts}~\citep{partee1995lexical}. Vanilla SAEs demonstrably succeed at many interpretability tasks, yet their linear reconstruction mechanism cannot, in principle, represent non-linear composition. Without explicit interaction mechanisms, SAEs cannot simultaneously represent atomic features and their non-linear compositions. When ``Starbucks'' appears in context, a linear model must either (i) allocate a dedicated feature for this compositional entity, sacrificing atomicity, or (ii) represent it through separate ``star'' and ``coffee'' features that cannot distinguish this specific composition from mere co-occurrence. 

Ideally, SAEs with sufficient capacity can learn features at multiple levels of abstraction simultaneously (such as morphemes, words, phrases, and compositional expressions) coexisting as independent atoms in an overcomplete dictionary. While this leads to good reconstruction and intervention, it fundamentally limits our understanding: we cannot decompose ``Starbucks'' into its constituents, cannot trace how ``administrators'' emerge from stem and suffix binding, and cannot distinguish compositional phrases from accidental co-occurrence. The conflation of atomic and compositional features \textit{obscures the mechanisms by which networks build complex representations from simpler parts}.

This problem connects to a \textit{longstanding debate} in cognitive science about systematic compositionality in neural representations~\citep{fodor1988connectionism}. \citet{smolensky1990tensor} proposed tensor product variable binding as a solution: \textit{features bind through multilinear interactions} rather than linear superposition, allowing networks to maintain atomic constituents while representing their combinations. In this framework, ``administrators'' would be represented not as a single indivisible feature, but as an explicit \textit{composition} of stem and suffix, where the tensor product captures the binding operation. For interpretability of modern LLMs, this principle is critical: to understand how networks compose meaning, our tools must themselves model compositional structure faithfully. However, explicit tensor products are computationally prohibitive for overcomplete sparse codes with tens of thousands of features, requiring methods that capture multilinear interactions while remaining tractable.

In this work, we introduce the \textbf{Polynomial Sparse Autoencoder (PolySAE)} (\Cref{fig:placeholder}), a sparse autoencoder that extends vanilla SAEs with explicit feature interaction terms. PolySAE preserves a linear encoder for interpretability while extending the decoder with quadratic and cubic terms that model pairwise and triple feature interactions. Through low-rank tensor factorization on a shared projection subspace, PolySAE captures compositional structure in a tractable manner, adding a small parameter overhead (3\% for GPT2 small). Critically, PolySAE is a strict \textit{generalization of standard SAEs} that enables capturing multiplicative (non-additive) concept interactions. Setting interaction coefficients to zero recovers vanilla SAE behavior, allowing PolySAE to be readily applied to existing SAE variants, including TopK \citep{gao2025scaling}, BatchTopK \citep{bussmann2024batchtopk}, and Matryoshka \citep{bussmann2025matryoshka}.

We summarize below our \textbf{four main contributions}:

\textbf{C1.} We introduce \textbf{PolySAE}, a sparse autoencoder with a polynomial decoder that explicitly \textbf{models quadratic and cubic feature interactions} while preserving a linear encoder for interpretability. Through low-rank tensor factorization, PolySAE adds small parameter overhead (3\% for GPT2 small) and can be \textbf{readily applied} to existing SAE variants (TopK, BatchTopK, Matryoshka).

\textbf{C2.} Across \textbf{four language models} of different scales (GPT-2 Small, Pythia-410M/1.4B, Gemma-2-2B) and \textbf{three sparsification strategies}, PolySAE achieves an \textbf{average 8\% F1 improvement}, while maintaining comparable reconstruction error.

\textbf{C3.} PolySAE produces \textbf{2--10$\times$ larger Wasserstein distances} between class-conditional feature distributions, indicating more separated semantic structure in the learned representations.

\textbf{C4.} We show that learned interaction weights exhibit \textbf{negligible correlation with co-occurrence frequency} (r = 0.06 vs.\ r = 0.82 for SAE feature covariance), and provide qualitative examples demonstrating that \textbf{polynomial terms capture compositional structure} such as morphological binding, phrasal composition, and contextual disambiguation. We further show that the learned interaction directions causally steer model outputs toward the corresponding compositional semantics.

%% file: sections/2.related_work.tex
\section{Related Work}\label{sec:related_work}
\textbf{Sparse dictionary learning.} In sparse dictionary learning, signals are represented as sparse linear combinations of overcomplete basis elements~\citep{mallat1993matching}, an approach also integrated into neural network architectures~\citep{hinton2006reducing,lee2007sparse,konda2014zero}. Sparse Autoencoders (SAEs) recently emerged as a leading paradigm for feature discovery in large language models~\citep{huben2024sparse,bricken2023monosemanticity}, scaling to millions of features~\citep{gao2025scaling}. Subsequent work has produced architectural variants including BatchTopK~\citep{bussmann2024batchtopk}, Matryoshka~\citep{bussmann2025matryoshka}, Gated~\citep{rajamanoharan2024improving}, and JumpReLU~\citep{rajamanoharan2024jumping} SAEs, with standardized benchmarks enabling systematic comparison~\citep{karvonen2025saebench}. However, all these methods assume features combine additively through linear reconstruction.

\textbf{Modeling feature interactions.} Multiplicative interactions between features have a rich history in deep learning~\citep{jayakumar2020multiplicative}, from early bilinear models for visual data~\citep{tenenbaum1996separating,freeman1997learning} to modern gating mechanisms~\citep{shazeer2020glu}. Feature interactions through the Hadamard product~\citep{chrysos2025hadamard} serve as a powerful conditioning mechanism~\citep{perez2018film,dumoulin2017a}, while multiplicative structure also enables parameter-efficient mixture-of-experts~\citep{oldfield2025towards}. Recent work has explored multiplicative interactions for interpretability: Bilinear MLPs~\citep{pearce2025bilinear} model pairwise feature interactions enabling weight-based interpretability, while \citet{gauderis2025compositionality} propose fully interpretable architectures based on tensor networks.  We extend this line of work by modeling feature interactions in the SAE setting.

\textbf{Polynomials}. One natural way to model higher-order interactions is through polynomials \citep{shin1991pi}. In deep learning, polynomials have been used for a variety of applications, such as image generation \citep{chrysos2020picvpr,chrysos2021conditional}, classification \cite{babiloni2021poly,chrysos2022augmenting}, privacy preservation \citep{zhang2019encrypted}, interpretability \citep{dubey2022scalable}, and dynamic safety guardrails \citep{oldfield2026beyond}.
The work most closely related to ours is the Bilinear Autoencoder (BAE) \citep{dooms2025bilinearauto}, which similarly introduces interaction terms for interpretability.
The key difference lies in the level at which interactions are modeled: BAE captures pairwise interactions between input neurons, whereas PolySAE models interactions directly between learned sparse features, including higher-order terms.
As a result, PolySAE preserves the interpretability of linear SAE latents while explicitly allocating capacity to non-additive feature composition.

%% file: sections/4.method.tex
\section{Sparse Polynomial Decoding}\label{sec:method}

\subsection{Preliminaries}
\label{sec:meth:preliminaries}

\textbf{Notation.}
Bold lowercase letters denote vectors and bold uppercase letters denote matrices.
The $i$-th column of $\mM$ is $\vm_i$, and $\mM_{:,1:r}$ denotes its first $r$ columns.
We use $*$ for the Hadamard product, $\otimes$ for the Kronecker product, and $\odot$ for the Khatri--Rao product.
$\sR^d$ and $\sR^{d_{\mathrm{sae}}}$ denote the activation and sparse-code spaces, with $d_{\mathrm{sae}} \gg d$.
$\cS(\cdot)$ denotes a sparsification operator, such as Top-$K$ or BatchTop-$K$.

\textbf{Sparse Autoencoders.}
Sparse autoencoders (SAEs) build on overcomplete dictionary learning~\citep{olshausen1997sparse} to decompose neural activations into a sparse set of latent features.
Given activations $\vx \in \sR^d$ from an intermediate layer of a pretrained network, an SAE learns a sparse code $\vz \in \sR^{d_{\mathrm{sae}}}$ with $d_{\mathrm{sae}} \gg d$ and reconstructs via $\hat{\vx} = \vb_{\mathrm{dec}} + \mD\vz, \vz = \cS\big(\mathrm{ReLU}(\mE^\top \vx + \vb_{\mathrm{enc}})\big)$ where $\mE$ is a linear encoder, $\mD$ is the decoder (dictionary), and $\cS$ enforces sparsity.
The overcomplete latent space allows multiple features to align with similar activation directions, supporting disentangled and interpretable representations.

Motivated by the superposition hypothesis~\citep{elhage2022toy}, SAEs assume that features combine \emph{additively} in the decoder, so reconstruction is linear in $\vz$.
This corresponds to a strong form of the linear representation hypothesis applied to decoding, which has recently been questioned~\citep{engelsnot}.
When multiple features co-activate, their joint effect may not be well captured by a linear sum—for example, a ``coffee'' feature and a ``star'' feature may require a reconstruction direction distinct from either individual atom to capture the ``Starbucks'' concept.
This motivates extending the decoder to explicitly model feature interactions, while preserving a linear and interpretable encoder.

\subsection{Design Principles for Feature Interactions}
\label{sec:meth:principles}

We extend sparse autoencoders to capture higher-order feature interactions by establishing design principles grounded in prior work. Each architectural choice in PolySAE follows directly from these principles.

\textbf{P1. Linear Encoding} (Interpretability). Each sparse code coefficient $z_i$ is derived by a \emph{linear projection} of the input activation $\vx$. The linear representation hypothesis in mechanistic interpretability posits that learned features should correspond to \emph{directions} in activation space~\citep{elhage2022toy,bricken2023monosemanticity}, a view supported by the success of linear probes for extracting semantic content~\citep{belinkov2022probing, alain2016understanding}.

\textbf{P2. Polynomial Reconstruction} (Expressivity). The decoder may capture compositional structure by using polynomial terms in $\vz$. Modeling \emph{compositional} structure, \ie how features interact, polynomials have a strong precedent in the literature: Volterra series~\citep{volterra1959theory} represent nonlinear systems as sums of multilinear kernels, second-order pooling~\citep{carreira2012semantic,gao2016compact} captures feature co-occurrences via outer products, and polynomial networks~\citep{chrysos2021deep} parameterize functions as products of linear projections.

\textbf{P3. Factorized Interaction Structure} (Coherence \& Efficiency). Higher-order terms should operate in a low-dimensional subspace aligned with the linear feature space. Using a shared projection $\mU$ ensures that interactions are compositions of the same underlying features. This alignment principle underlies factorized interaction models~\citep{rendle2010factorization, blondel2016higher} and compact bilinear pooling~\citep{gao2016compact, kim2017hadamard}. Constraining interactions to low-rank subspaces imposes a strong inductive bias, favoring coherent, reusable interaction modes over arbitrary pairwise composition.

\textbf{P4. Structural Constraints} (Parsimony \& Identifiability). Lower-order terms should have higher representational capacity than higher-order terms, following polynomial approximation theory~\citep{mason2002chebyshev}. The latent interaction subspace should have orthonormal columns to ensure geometrically distinct directions. Orthogonality constraints are standard in dictionary learning and independent component analysis to prevent degenerate solutions~\citep{arora2015simple,bao2016dictionary,hyvarinen2000independent}. Orthonormality removes rotational ambiguity and ensures the model does not allocate redundant capacity to correlated interaction directions.

\subsection{PolySAE: Polynomial Sparse Autoencoder}
\label{sec:meth:architecture}

To satisfy \textbf{P1} PolySAE adopts the standard SAE encoder~\citep{huben2024sparse,bricken2023monosemanticity} to first performs a linear map followed by sparsification:
\begin{equation}
\vz = \mathcal{S}\big(\mathrm{ReLU}(\mE^\top \vx + \vb_{\mathrm{enc}})\big),
\qquad
\vz \in \sR^{d_{\mathrm{sae}}},
\label{eq:encoder}
\end{equation}
where feature $i$ activates when $\vx$ aligns with direction $\ve_i$, enabling visualization, clustering, and causal intervention via activation patching~\citep{meng2022locating}. 

\looseness-1Following \textbf{P2}, we extend the decoder to include quadratic and cubic terms:
\begin{equation}
\hat{\vx} = \vb_{\mathrm{dec}} + \vy_1 + \lambda_2\,\vy_2 + \lambda_3\,\vy_3,
\label{eq:poly_decoder}
\end{equation}
where $\vy_1 = \mA\,\vz$, $\vy_2 = \mB\,(\vz\otimes\vz)$, $\vy_3 = \mGamma\,(\vz\otimes\vz\otimes\vz)$, and $\lambda_2, \lambda_3 \in \R$ are learnable scalar coefficients that control the contribution of each polynomial order. Setting $\lambda_2 = \lambda_3 = 0$ recovers a standard linear sparse autoencoder, making PolySAE a strict generalization of existing SAE architectures. This can be viewed as a third-order Volterra expansion~\citep{volterra1959theory} or a \emph{$\Pi$-net} polynomial parameterization~\citep{chrysos2021deep}, adapted to sparse codes.

However, explicitly modeling all pairwise or higher-order feature combinations would require $O(d_{\mathrm{sae}}^2)$ or $O(d_{\mathrm{sae}}^3)$ parameters, leading to unstructured interaction effects and a high risk of overfitting. Following \textbf{P3} and \textbf{P4}, we constrain interactions to a low-rank subspace:
\begin{equation}
\begin{aligned}
\vy_{1} &= (\vz\,\mU)\,\mC^{(1)\!\top},\\[2mm]
\vy_{2} &= \big((\vz\,\mU_{:,1:R_2}) * (\vz\,\mU_{:,1:R_2})\big)\,\mC^{(2)\!\top},\\[2mm]
\vy_{3} &= \big((\vz\,\mU_{:,1:R_3}) * (\vz\,\mU_{:,1:R_3}) * (\vz\,\mU_{:,1:R_3})\big)\,\mC^{(3)\!\top},
\end{aligned}
\label{eq:lowrank_decoder}
\end{equation}
where $*$ denotes element-wise product and $\mC^{(k)} \in \sR^{d \times R_k}$ are output projection matrices. This parameterization restricts the  interaction dictionaries to rank at most $R_k$, enforcing a strong inductive bias on how features may combine.

Notice that this parameterization satisfies \textbf{P3} by applying a single projection $\mU$ to the sparse code and forming interactions via polynomial operations: $\vz\mU$, $(\vz\mU) * (\vz\mU)$, and $(\vz\mU) * (\vz\mU) * (\vz\mU)$. Using the same projected representation at every order ensures that interaction effects remain aligned with the linear feature basis and interpretable as compositions of the same underlying features. 

Furthermore, PolySAE satisfies the parsimony aspect of \textbf{P4} by following nested low-rank approximation~\citep{grasedyck2013literature} and utilizing ranks $(R_1, R_2, R_3)$ with $R_1 \geq R_2 \geq R_3$. This nested structure means $\mathrm{span}(\mU_{:,1:R_3}) \subset \mathrm{span}(\mU_{:,1:R_2}) \subset \mathrm{span}(\mU)$. In practice, $R_2 = R_3 \ll R_1$ (\eg, $R_2 = R_3 = 64$) suffices to capture most interaction structure, confirming our hypothesis that higher-order contributions are low-dimensional (\cref{sec:experiments}).

Finally, to satisfy the identifiability aspect of \textbf{P4}, we enforce orthonormality of the interaction subspace. Following Stiefel optimization~\citep{absil2008optimization,bonnabel2013stochastic}, we impose $\mU^\top \mU=\mI$ via QR retraction after each gradient step. We use positive QR retraction~\citep{edelman1998geometry}, which corrects column signs to ensure continuity and avoids discontinuous representation changes during training.

\subsection{Discussion}
\label{sec:meth:discussion}

\textbf{Context-Dependent Dictionary Structure.}
In standard SAEs, each feature $i$ is associated with a fixed dictionary atom $\vd_i$: regardless of context, activating feature $i$ contributes $z_i \vd_i$ to the reconstruction.
PolySAE fundamentally alters this picture. Because reconstruction includes higher-order terms, the effective contribution of a feature becomes \emph{context-dependent}, varying with which other features are simultaneously active.

This can be seen by expanding \Cref{eq:poly_decoder}. The linear term defines a dictionary over individual features, while the quadratic and cubic terms define dictionaries over feature pairs and triples, respectively. Under our low-rank factorization, these dictionaries are implicitly given by
\begin{equation}
\begin{aligned}
\mA &\;=\; \mC^{(1)}\,\mU^{\top} &&\in\; \sR^{d \times d_{\mathrm{sae}}},\\[2mm]
\mB &\;=\; \mC^{(2)}\!\left(\mU_{:,1:R_2} \odot \mU_{:,1:R_2}\right)^{\!\top} &&\in\; \sR^{d \times d_{\mathrm{sae}}^{2}},\\[2mm]
\mGamma &\;=\; \mC^{(3)}\!\left(\mU_{:,1:R_3} \odot \mU_{:,1:R_3} \odot \mU_{:,1:R_3}\right)^{\!\top} &&\in\; \sR^{d \times d_{\mathrm{sae}}^{3}},
\end{aligned}
\label{eq:polysae_dictionaries}
\end{equation}
where $\mA$ is the \emph{linear dictionary}, $\mB$ the \emph{pairwise interaction dictionary}, and $\mGamma$ the \emph{triple interaction dictionary}. Column $(i,j)$ of $\mB$ specifies how the co-activation $z_i z_j$ modifies the reconstruction, while column $(i,j,k)$ of $\mGamma$ specifies the contribution arising from the joint activation $z_i z_j z_k$. The computational form in \Cref{eq:lowrank_decoder} is algebraically equivalent to \Cref{eq:polysae_dictionaries} but avoids explicitly materializing the $d_{\mathrm{sae}}^2$- and $d_{\mathrm{sae}}^3$-dimensional dictionaries.

\textbf{Compositional Capacity.}
Using the same $d_{\mathrm{sae}}$ base features as an SAE, PolySAE can support interaction-driven structure across
$\binom{d_{\mathrm{sae}}}{2} \cdot R_2
\;+\;
\binom{d_{\mathrm{sae}}}{3} \cdot R_3$
feature pairs and triples, enabling a substantially larger space of distinct semantic compositions without increasing the number of features. This capacity is mediated through a shared low-rank interaction space: rather than allocating independent parameters to each feature combination, interactions are expressed via $R_2$ and $R_3$ shared modes. As a result, potential feature combinations are realized through a small number of reusable interaction directions, reflecting the empirically observed low-dimensional structure of feature interactions.

\textbf{Parameter Efficiency.}
PolySAE modifies only the decoder; the encoder is unchanged.
A standard SAE has $2d\,d_{\mathrm{sae}} + d + d_{\mathrm{sae}}$ parameters.
When the linear term is full rank ($R_1=d$), PolySAE adds $\Delta P \;=\; d^2 + d(R_2+R_3) + 2$ parameters. With the empirically optimal choice $R_2=R_3$ and $R_2 \in [0.06R_1,\,0.11R_1]$, this yields
$\Delta P = (1.12\text{--}1.22)\,d^2$ (up to constants).
For GPT-2 small ($d=768$, $d_{\mathrm{sae}}=16{,}384$), this corresponds to an increase of $\sim\!2.5$–$3\%$ of the full SAE.

%% file: sections/5.experiments.tex
\begin{table*}[h]
\caption{F1 Scores (\%) across datasets at K=1. Format: F1 / Wasserstein ($\times 10^{-3}$). Mean Probing column shows mean F1 across datasets. MSE for reconstruction error and CE Rec.\ denotes cross-entropy recovery.}
\label{tab:f1_scores_k1}
\centering
\scriptsize
\renewcommand{\arraystretch}{0.9} 
\setlength{\tabcolsep}{4pt}
\renewcommand{\arraystretch}{1.15}
\begin{tabular}{clccc|cccccc}
\toprule
LLM & SAE variant & MSE & CE Rec.\ & Mean F1 & Europarl & Bios & \makecell{Amazon \\ Sentiment} & GitHub & AG News & Amazon 15 \\
\midrule
\multirow{6}{*}{GPT-2 Small}
  & Topk &\bfseries 0.52 &\bfseries 0.993 & 67.1 & 67.7 / 19.0 & 61.0 / 7.7 & 76.0 / 4.3 & 63.4 / 8.7 & 71.4 / 8.5 & 63.3 / 2.8 \\
  &\cellcolor{cyan!5}Topk + PolySAE &\cellcolor{cyan!5} 0.55 &\cellcolor{cyan!5}\bfseries 0.993 &\cellcolor{cyan!5}\bfseries 77.9 &\cellcolor{cyan!5}\bfseries 86.1 / 35.2 &\cellcolor{cyan!5}\bfseries 75.5 / 16.8 &\cellcolor{cyan!5}\bfseries 83.1 / 9.7 &\cellcolor{cyan!5}\bfseries 73.0 / 20.6 &\cellcolor{cyan!5}\bfseries 81.0 / 18.9 &\cellcolor{cyan!5}\bfseries 69.0 / 6.7 \\ \cline{2-11}
  & BTopk &\bfseries 0.53 &\bfseries 0.993 & 65.7 & 67.4 / 19.0 & 59.6 / 7.3 & 68.8 / 4.4 & 68.1 / 8.8 & 65.3 / 8.2 & 65.1 / 2.9 \\
  &\cellcolor{cyan!5}BTopk + PolySAE &\cellcolor{cyan!5} 0.54 &\cellcolor{cyan!5}\bfseries 0.993 &\cellcolor{cyan!5}\bfseries 78.0 &\cellcolor{cyan!5}\bfseries 92.0 / 39.9 &\cellcolor{cyan!5}\bfseries 70.9 / 17.3 &\cellcolor{cyan!5}\bfseries 84.2 / 8.5 &\cellcolor{cyan!5}\bfseries 74.4 / 18.5 &\cellcolor{cyan!5}\bfseries 83.2 / 20.0 &\cellcolor{cyan!5}\bfseries 63.2 / 6.0 \\ \cline{2-11}
  & Matryoshka & 0.60 &\bfseries 0.992 & 65.7 & 65.8 / 12.5 & 61.2 / 4.0 & 76.2 / 3.2 & 60.9 / 7.9 & 68.1 / 4.3 & 62.1 / 2.4 \\
  &\cellcolor{cyan!5}Matr.\ + PolySAE &\cellcolor{cyan!5}\bfseries 0.58 &\cellcolor{cyan!5}\bfseries 0.992 &\cellcolor{cyan!5}\bfseries 77.7 &\cellcolor{cyan!5}\bfseries 95.0 / 30.0 &\cellcolor{cyan!5}\bfseries 72.9 / 14.3 &\cellcolor{cyan!5}\bfseries 81.4 / 8.1 &\cellcolor{cyan!5}\bfseries 71.5 / 18.6 &\cellcolor{cyan!5}\bfseries 77.4 / 16.0 &\cellcolor{cyan!5}\bfseries 68.0 / 5.6 \\
\midrule
\multirow{6}{*}{Pythia-410m}
  & Topk &\bfseries 0.03 &\bfseries 0.971 & 71.2 & 96.1 / 2.0 & 67.4 / 1.1 & 61.5 / 0.7 & 64.6 / 1.5 & 71.8 / 1.2 & 65.9 / 0.4 \\
  &\cellcolor{cyan!5}Topk + PolySAE &\cellcolor{cyan!5} 0.04 &\cellcolor{cyan!5} 0.970 &\cellcolor{cyan!5}\bfseries 77.0 &\cellcolor{cyan!5}\bfseries 96.7 / 6.8 &\cellcolor{cyan!5}\bfseries 70.8 / 3.8 &\cellcolor{cyan!5}\bfseries 75.9 / 2.3 &\cellcolor{cyan!5}\bfseries 74.0 / 5.3 &\cellcolor{cyan!5}\bfseries 73.3 / 4.0 &\cellcolor{cyan!5}\bfseries 71.5 / 1.4 \\ \cline{2-11}
  & BTopk &\bfseries 0.03 &\bfseries 0.973 & 65.0 & 90.9 / 0.8 & 60.5 / 0.3 & 63.9 / 0.4 & 59.7 / 1.1 & 58.7 / 0.3 & 56.6 / 0.3 \\
  &\cellcolor{cyan!5}BTopk + PolySAE &\cellcolor{cyan!5} 0.04 &\cellcolor{cyan!5} 0.971 &\cellcolor{cyan!5}\bfseries 77.3 &\cellcolor{cyan!5}\bfseries 97.8 / 8.2 &\cellcolor{cyan!5}\bfseries 74.0 / 4.1 &\cellcolor{cyan!5}\bfseries 74.6 / 2.1 &\cellcolor{cyan!5}\bfseries 75.2 / 4.8 &\cellcolor{cyan!5}\bfseries 78.6 / 4.4 &\cellcolor{cyan!5}\bfseries 63.5 / 1.3 \\ \cline{2-11}
  & Matryoshka &\bfseries 0.04 & 0.969 & 64.2 & 79.1 / 0.6 & 63.6 / 0.3 & 64.4 / 0.4 & 62.3 / 1.1 & 58.7 / 0.3 & 57.3 / 0.3 \\
  &\cellcolor{cyan!5}Matr.\ + PolySAE &\cellcolor{cyan!5}\bfseries 0.04 &\cellcolor{cyan!5}\bfseries 0.972 &\cellcolor{cyan!5}\bfseries 74.6 &\cellcolor{cyan!5}\bfseries 99.2 / 2.8 &\cellcolor{cyan!5}\bfseries 71.0 / 1.2 &\cellcolor{cyan!5}\bfseries 66.9 / 1.3 &\cellcolor{cyan!5}\bfseries 81.8 / 3.7 &\cellcolor{cyan!5}\bfseries 64.8 / 1.2 &\cellcolor{cyan!5}\bfseries 63.8 / 0.9 \\
\midrule
\multirow{6}{*}{Pythia-1.4b}
  & Topk &\bfseries 0.23 & 0.971 & 75.9 & \textbf{97.8} / 1.6 & 72.5 / 1.4 & 69.5 / 0.8 & 69.3 / 1.9 & 77.3 / 1.4 & 69.0 / 0.5 \\
  &\cellcolor{cyan!5}Topk + PolySAE &\cellcolor{cyan!5}\bfseries 0.23 &\cellcolor{cyan!5}\bfseries 0.973 &\cellcolor{cyan!5}\bfseries 81.9 &\cellcolor{cyan!5} 96.8 / \textbf{7.9} &\cellcolor{cyan!5}\bfseries 77.2 / 6.4 &\cellcolor{cyan!5}\bfseries 88.1 / 3.7 &\cellcolor{cyan!5}\bfseries 74.7 / 9.2 &\cellcolor{cyan!5}\bfseries 83.4 / 6.3 &\cellcolor{cyan!5}\bfseries 71.1 / 2.4 \\ \cline{2-11}
  & BTopk &\bfseries 0.22 &\bfseries 0.975 & 64.6 & 74.0 / 0.6 & 65.0 / 0.5 & 57.2 / 0.6 & 63.3 / 2.3 & 65.2 / 0.4 & 63.2 / 0.4 \\
  &\cellcolor{cyan!5}BTopk + PolySAE &\cellcolor{cyan!5} 0.23 &\cellcolor{cyan!5} 0.974 &\cellcolor{cyan!5}\bfseries 76.4 &\cellcolor{cyan!5}\bfseries 93.7 / 4.5 &\cellcolor{cyan!5}\bfseries 73.0 / 3.4 &\cellcolor{cyan!5}\bfseries 67.1 / 3.1 &\cellcolor{cyan!5}\bfseries 73.8 / 8.2 &\cellcolor{cyan!5}\bfseries 77.6 / 3.4 &\cellcolor{cyan!5}\bfseries 73.2 / 2.1 \\ \cline{2-11}
  & Matryoshka & 0.24 & 0.970 & 64.4 & 70.2 / 0.5 & 62.8 / 0.5 & \textbf{63.1} / 0.6 & 65.6 / 1.9 & 64.1 / 0.4 & 60.8 / 0.4 \\
  &\cellcolor{cyan!5}Matr.\ + PolySAE &\cellcolor{cyan!5}\bfseries 0.23 &\cellcolor{cyan!5}\bfseries 0.973 &\cellcolor{cyan!5}\bfseries 72.1 &\cellcolor{cyan!5}\bfseries 91.1 / 2.9 &\cellcolor{cyan!5}\bfseries 72.4 / 2.0 &\cellcolor{cyan!5} 58.0 / \textbf{2.1} &\cellcolor{cyan!5}\bfseries 68.2 / 6.7 &\cellcolor{cyan!5}\bfseries 73.6 / 2.1 &\cellcolor{cyan!5}\bfseries 69.4 / 1.5 \\
\midrule
\multirow{6}{*}{Gemma2-2b}
  & Topk &\bfseries 1.59 &\bfseries 0.988 & 67.7 & 78.6 / 5.3 & \textbf{69.6} / 7.1 &\textbf{71.8} / 4.4 & 60.7 / 6.1 & 60.7 / 7.2 & 64.8 / 2.7 \\
  &\cellcolor{cyan!5}Topk + PolySAE &\cellcolor{cyan!5} 1.65 &\cellcolor{cyan!5} 0.987 &\cellcolor{cyan!5}\bfseries 68.4 &\cellcolor{cyan!5}\bfseries 86.8 / 12.0 &\cellcolor{cyan!5} 64.7 / \textbf{16.8} &\cellcolor{cyan!5} 64.5 / \textbf{10.5} &\cellcolor{cyan!5}\bfseries 64.1 / 16.1 &\cellcolor{cyan!5}\bfseries 61.9 / 16.9 &\cellcolor{cyan!5}\bfseries 68.5 / 6.3 \\ \cline{2-11}
  & BTopk &\bfseries 1.58 &\bfseries 0.988 & 64.8 & 68.3 / 1.9 & 67.6 / 2.6 & \textbf{71.1} / 2.8 & 64.4 / 4.5 & 59.9 / 2.7 & 57.6 / 1.9 \\
  &\cellcolor{cyan!5}BTopk + PolySAE &\cellcolor{cyan!5} 1.68 &\cellcolor{cyan!5} 0.987 &\cellcolor{cyan!5}\bfseries 69.4 &\cellcolor{cyan!5}\bfseries 92.8 / 13.2 &\cellcolor{cyan!5}\bfseries 78.3 / 18.3 &\cellcolor{cyan!5} 56.4 / \textbf{10.2} &\cellcolor{cyan!5}\bfseries 65.0 / 16.1 &\cellcolor{cyan!5}\bfseries 64.0 / 18.8 &\cellcolor{cyan!5}\bfseries 60.0 / 6.4 \\ \cline{2-11}
  & Matryoshka & 1.69 &\bfseries 0.987 & 60.9 & 60.8 / 0.7 & 64.0 / 0.8 & 57.3 / 1.5 & 61.5 / 2.5 & \textbf{61.0} / 0.8 & 60.5 / 1.0 \\
  &\cellcolor{cyan!5}Matr.\ + PolySAE &\cellcolor{cyan!5}\bfseries 1.64 &\cellcolor{cyan!5}\bfseries 0.987 &\cellcolor{cyan!5}\bfseries 65.6 &\cellcolor{cyan!5}\bfseries 77.6 / 2.1 &\cellcolor{cyan!5}\bfseries 67.5 / 3.1 &\cellcolor{cyan!5}\bfseries 61.7 / 4.9 &\cellcolor{cyan!5}\bfseries 63.7 / 8.8 &\cellcolor{cyan!5} 60.9 / \textbf{3.5} &\cellcolor{cyan!5}\bfseries 62.2 / 3.3 \\
\bottomrule
\end{tabular}
\end{table*}

\section{Empirical Evaluation}\label{sec:experiments}

\subsection{Experimental Setup}
\label{sec:main_setup}

Our training pipeline is built by extending \texttt{SAELens} \citep{bloom2024saetrainingcodebase} to include PolySAE.
We train and evaluate our methods against the standard SAE with \textbf{three sparsification strategies}, TopK \cite{gao2025scaling}, BatchTopK \cite{bussmann2024batchtopk}, and Matryoshka \cite{bussmann2025matryoshka}.
Throughout all experiments, we use a sparsity level of $K=64$ with $16{,}384$ latents trained on residual-stream activations from \textbf{four pretrained language models of different scales}: Gemma-2-2B \citep{gemma2report} (layer 19),
Pythia-410M and Pythia-1.4B \citep{biderman2023pythia} (layers 15 and 12, respectively),
and GPT-2 Small \citep{radford2019gpt2} (layer 8).
Training uses 500M tokens (300M for GPT-2 Small) with context length 128.
For Gemma-2-2B and GPT-2 Small, we use OpenWebText \citep{gokaslan2019openwebtext};
for Pythia models, we use an uncopyrighted variant of the deduplicated Pile \citep{gao2021pile}.
We evaluate learned features using SAEBench \citep{karvonen2025saebench}, which reports reconstruction metrics on held-out data from the training distribution and sparse probing performance on \textbf{six classification tasks}: Bias in Bios \citep{dearteaga2019biasinbios}, AG News \citep{zhang2015charcnn}, EuroParl \citep{koehn-2005-europarl}, GitHub programming languages \citep{codeparrot_github_code}, Amazon Sentiment, and Amazon-15 \citep{hou2024bridging}. For more implementation details see \Cref{sec:appendix_implementation}.

\vspace{-4.4pt}

\subsection{Reconstruction and Semantic Modeling}
\label{sec:exp:rec_probing_metrics}

We evaluate models along two axes: \textit{(Q1) reconstruction fidelity} and \textit{(Q2) semantic modeling of the learned representations}.
Reconstruction quality is measured using mean squared error between the decoder output and the unnormalized network activations.
To assess semantic structure, we use two complementary metrics.

\textit{Probing.}
We evaluate the linear separability of semantic concepts in the learned sparse representations by training logistic regression classifiers on SAE activations to predict ground-truth labels across multiple datasets.
For each task, classification is performed using the feature with the largest mean activation difference between positive and negative classes, isolating semantic signal at the feature level.

\textit{Distributional separation.}
Probing relies on post-hoc decision boundaries and may not fully reflect the intrinsic geometry of the representation.
We therefore additionally compute the 1-Wasserstein distance between class-conditional activation distributions.
Unlike probing, which evaluates separability at a specific threshold, the Wasserstein distance captures global distributional separation, with larger values indicating more distinct semantic separation across space.

\Cref{tab:f1_scores_k1} demonstrates that \textbf{across four language models and three sparsification strategies}, PolySAE achieves \textit{comparable MSE to standard SAE} across all configurations, confirming that polynomial decoding does not sacrifice reconstruction fidelity. For probing, PolySAE \textbf{consistently outperforms SAE by large margins} with mean gains of more than 10\% on GPT-2, and 8\% on average across models (Pythia-410M, Pythia-1.4B, and Gemma2-2B) and sparsifiers. Crucially, PolySAE also achieves   \textbf{consistently substantially higher Wasserstein distances}, with improvements of approximately $2$--$10\times$ across all other models. This indicates that the gains observed in probing accuracy reflect \textit{genuinely better-separated class-conditional representations}, rather than improvements driven solely by favorable decision boundaries. We additionally report cross-entropy (CE) loss recovery in \Cref{tab:f1_scores_k1}, which captures whether the SAE preserves model behavior when its reconstruction substitutes for the original activation. Across all 12 configurations PolySAE's CE recovery lies within 0.003 of vanilla SAE, confirming that the small MSE differences do not translate into functional degradation.

\subsection{Ablations}

\textbf{Ablating P3 and P4.}
We next ask \textit{(Q3) whether each architectural choice in PolySAE carries its weight}. Starting from a polynomial decoder with full-rank shared projections, we additively introduce low-rank factorization (P3) and orthogonality (P4). \Cref{tab:arch_ablation} reports parameters, MSE, and F1 on GPT-2 Small. Low-rank factorization cuts parameters by \textit{65\% at only 1pp F1 cost}, making training tractable. Orthogonality then \textit{recovers the gap and surpasses the original polynomial by +2.9pp at zero parameter cost}, validating P4.

\textbf{PolySAE Enables Competitive Performance with Sparser Codes.}
We ask \textit{(Q4) whether PolySAE's capacity to model feature interactions enables the use of sparser representations}.
\Cref{fig:sparsity} shows probing F1 as a function of active features $k$, with shaded regions indicating variance across dictionary widths (2k--16k).
PolySAE consistently outperforms standard SAEs across all sparsity levels, with the gap widening at higher $k$.
PolySAE also exhibits lower variance across widths, enabling competitive performance with fewer active features.

\begin{table}[]
\caption{Architectural ablation on GPT-2 Small validating P3 and P4.}
\label{tab:arch_ablation}
\centering
\footnotesize
\setlength{\tabcolsep}{4pt}
\renewcommand{\arraystretch}{1.15}
\begin{tabular}{l c c c}
\toprule
Ablation & Params & MSE & F1 \\
\midrule
Polynomial + shared proj.\        & 37.7M & 0.58 & 76.0 \\
\quad + low-rank factorization     & 13.3M & \bfseries 0.53 & 75.0 \\
\quad + orthogonality (PolySAE)    & 13.3M & 0.55 & \bfseries 77.9 \\
\bottomrule
\end{tabular}
\end{table}

\input{figures/topk_width}

\textbf{Semantic Concentration Across Features.} We further ask \textit{(Q5) whether PolySAE concentrates semantic signal into fewer features}.
\Cref{tab:f1_gain_by_model} reports the F1 gain $\Delta_{1\text{--}5}$ when expanding from K=1 to K=5 active features, broken down by model and sparsifier across the probing datasets.
PolySAE exhibits smaller gains than standard SAEs in 9 out of 12 model$\times$sparsifier configurations, with all three GPT-2 Small sparsifiers showing gaps of $-7$ to $-8$.
This behavior is probably due to the fact that higher-order interactions absorb contextual variability while PolySAE's linear features remain more semantically focused.

\begin{table}[]
\caption{Mean F1 Gain from K=1 to K=5 per model and sparsifier, averaged across the 6 probing datasets. \textbf{Bold} marks the lower (better) value per row, indicating stronger semantic concentration into fewer features.}
\label{tab:f1_gain_by_model}
\centering
\footnotesize
\renewcommand{\arraystretch}{1.15}
\begin{tabular}{l l c c}
\toprule
LLM & Sparsifier & SAE & PolySAE \\
\midrule
GPT-2 Small & TopK       & +15.2 & \bfseries +6.8 \\
GPT-2 Small & BatchTopK  & +13.6 & \bfseries +6.5 \\
GPT-2 Small & Matryoshka & +14.2 & \bfseries +6.8 \\
\midrule
Pythia-410m & TopK       & +10.2 & \bfseries +9.8  \\
Pythia-410m & BatchTopK  & +12.1 & \bfseries +11.2 \\
Pythia-410m & Matryoshka & +10.7 & \bfseries +10.4 \\
\midrule
Pythia-1.4b & TopK       & +10.2          & \bfseries +6.6  \\
Pythia-1.4b & BatchTopK  & \bfseries +9.7 & +10.6 \\
Pythia-1.4b & Matryoshka & \bfseries +11.1 & +14.3 \\
\midrule
Gemma2-2b & TopK       & +16.9          & \bfseries +9.2  \\
Gemma2-2b & BatchTopK  & \bfseries +12.6 & +12.8 \\
Gemma2-2b & Matryoshka & +11.2          & \bfseries +9.7  \\
\bottomrule
\end{tabular}
\end{table}

\begin{table*}[t]
\centering
\footnotesize
\setlength{\tabcolsep}{1pt}
\renewcommand{\arraystretch}{1.05}
\setlength{\emergencystretch}{2em}
\sloppy
\caption{\textbf{Second-Order Interaction Examples Captured by PolySAE.}
\looseness-1Quadratic interactions bind two features to capture context-dependent semantic structure beyond  co-occurrence.
SAE often recovers individual components but fails to represent the composed meaning.}
\label{tab:polysae_second_order}
\begin{tabularx}{\textwidth}{@{}L{1.4cm} L{1.5cm} @{\hspace{0.2cm}} Y @{\hspace{0.2cm}} L{1.6cm} Y@{}}
\toprule
\textbf{Poly $F_1$} &
\textbf{Poly $F_2$} &
\textbf{Context} &
\textbf{SAE}&
\textbf{Observed Pattern} \\
\midrule
{[star, stars]} &
{[coffee, tea]} &
We've all certainly heard of beers brewed with espresso, but how about one with an espresso shot poured over the top? \textbf{Starbucks} &
{[Apple, Google]} &{\textit{\small The interaction binds features to represent a specific named entity creating a new semantic category.
}} \\
\addlinespace\relax
{[surgery, repair]} &
{[Trans, LGBT]} &
Some in the transgender community are worried a suspicious fire at a Montreal clinic will add delays to an already lengthy process to get gender reassignment \textbf{surgery} &
{[birth, baby]} &{\textit{\small Specialization: a general concept (surgery) gets specialized by domain context (Trans,LGBT) narrowing the semantic scope.
}} \\
\addlinespace\relax
{[DNA, genetic]} &
{[mod, mods]} &
Activists are opening up a new front in their campaign against genetic modification. The latest target is genetically-\textbf{mod}ified trees &
{[modified, edit]} & {\textit{\small Multiple modifiers stack to  create specific compound meanings. Interaction binds {\it genetic} with the action {\it modification}.
}} \\
\addlinespace\relax
{[secret, hidden]} &
{[Snowden, WikiLeaks]} &
On May 24th PBS aired a Frontline documentary about alleged Wikileaker Bradley Manning called ``\textbf{WikiSecrets}'' &
{[secret, secrets]} & {\textit{\small Feature interaction binds topical concepts to create coined term that could not be modeled via co-occurrence alone.
}} \\
\bottomrule
\end{tabularx}
\end{table*}

\begin{table*}[!t]
\centering
\footnotesize
\setlength{\tabcolsep}{1pt}
\renewcommand{\arraystretch}{1.05}
\setlength{\emergencystretch}{2em}
\sloppy
\caption{\textbf{Third-Order Interaction Examples Captured by PolySAE.}
Cubic interactions condition pairwise compositions on additional context, disambiguating meaning through three-way binding. Vanilla SAE typically activates broader or less specific features.}
\label{tab:polysae_third_order}
\begin{tabularx}{\textwidth}{@{}L{1.5cm} L{1.5cm}@{\hspace{0.2cm}} L{1.5cm} Y L{1.5cm}@{\hspace{0.2cm}} Y@{}}
\toprule
\textbf{Poly $F_1$} &
\textbf{Poly $F_2$} &
\textbf{Poly $F_3$} &
\textbf{Context} &
\textbf{SAE}&
\textbf{Observed pattern} \\
\midrule
{[proved, proven]} &
{[star, stars, superstar]} &
{[reputation, fame]} &
David Bowie proved some \textbf{stars} are big enough not to make themselves available &
{[star, stars, superstar]} &{\textit{\small Three-way relational binding, all  arguments must be present; reputation disambiguates which aspect of stars is relevant to the proving action.
}} \\
\addlinespace\relax
{[nuclear, reactor]} &
{[test, testing]} &
{[radiation, magnetic]} &
US tests \textbf{nuclear}-capable missile with the range to strike North Korea &
{[nuclear, atomic]} & {\textit{\small Specifying concept; Event type (testing) $\times$ domain (nuclear) $\times$ capability (radiation) \citep{parsons1990events}
}} \\
\addlinespace\relax
{[black, racial]} &
{[Americans, Canadians]} &
{[people, women]} &
In a push to get more Black \textbf{Americans} involved in the world of tech &
{[Americans, Muslims, Jews]} & {\textit{\small  Multi-attribute category intersection, binding demographic attributes.
}} \\
\addlinespace\relax
{[ing, ting]} &
{[stock, market]} &
{[invest, investing]} &
\textbf{Invest}ing.com --- Philippines stocks were higher after &
{[ing, training, running]} & {\textit{\small Three-way interaction between morphological marker (ing) and domain (stock, market) \TODO{shifts the concept invest from action to entity}\citep{asher2011lexical}.
}} \\
\bottomrule
\end{tabularx}
\end{table*}

\begin{table*}[t]
\centering
\caption{Selected examples of activation steering on GPT-2 Small. We add the first-order decoder directions $\mathbf{d}_i + \mathbf{d}_j$ to the residual stream (layer~8) during greedy generation. Each row shows the continuation under three conditions: no steering, vanilla SAE, and PolySAE. \textbf{Bold} marks where PolySAE steers generation toward the compositional target. The full set of examples is in \Cref{tab:steering_examples_full}.}
\label{tab:steering_examples_main}
\small
\setlength{\tabcolsep}{4pt}
\begin{tabular}{p{2.8cm} p{2.4cm} p{3.1cm} p{3.1cm} p{3.1cm}}
\toprule
\textbf{Features $\to$ Target} & \textbf{Prompt} & \textbf{No Steering} & \textbf{SAE} & \textbf{PolySAE} \\
\midrule

\scriptsize\texttt{[surgery]}$\times$\texttt{[trans]} \newline $\to$ \textit{gender}
& ``The procedure that helps individuals align their body with their identity is''
& called ``\textit{body alignment}.'' The procedure involves the body aligning$\ldots$
& called ``\textit{body alignment}.'' The procedure involves the use of a combination$\ldots$
& called ``\textbf{\textit{gender identity surgery}}.'' Performed by a surgeon who specializes in gender$\ldots$ \\
\midrule

\scriptsize\texttt{[canada]}$\times$\texttt{[oil]} \newline $\to$ \textit{Keystone}
& ``The controversial cross-border pipeline project is called the''
& \textit{Trans-Pacific Partnership} (TPP), a major step forward for the U.S.\ and Canada.
& \textit{Trans Mountain pipeline}, a controversial project in the works for years.
& \textbf{\textit{Keystone XL}}. The pipeline would carry crude oil from Alberta to U.S.\ refineries. \\
\midrule

\scriptsize\texttt{[involved]}$\times$\texttt{[support]} \newline $\to$ \textit{community}
& ``The foundation wants more people to become''
& \textit{entrepreneurs}, and it wants to make sure they're not just part of the problem.
& \textit{entrepreneurs}, and it wants to make sure they're not just part of the problem.
& \textbf{\textit{involved in the community}}, and to help them make a difference. \\
\midrule

\scriptsize\texttt{[economic]}$\times$\texttt{[times]} \newline $\to$ \textit{Economist}
& ``The magazine with coverage of world politics and business is The''
& \textit{New York Times}. The magazine with coverage of world politics and business is The New York Times.
& \textit{New York Times}. The New York Times is a daily newspaper in the United States$\ldots$
& \textbf{\textit{Economist}}. The Economist is a magazine that is a global news magazine$\ldots$ \\

\bottomrule
\end{tabular}
\end{table*}

\section{Understanding and Utilizing Interactions}

\subsection{Making Sense of Learned Interactions}
\looseness-1To better understand the learnt interactions, we firstly ask \textit{(Q6) whether PolySAE’s higher-order terms encode genuine compositional structure or merely reflect surface-level co-occurrence}.
\looseness-1To study this, we analyze SAE and PolySAE activations trained with Top-$K$ sparsification on GPT-2 small, using 1M OpenWebText texts. For each feature pair $(i,j)$, we first define the learned quadratic interaction strength $B_{ij} = \lambda_2 \left\lVert (\vu_i \odot \vu_j)^\top \mC^{(2)} \right\rVert_2,$ which depends only on the trained decoder parameters and measures how much representational capacity the model allocates to the $(i,j)$ interaction.
Second, we compute the empirical co-occurrence frequency $N_{ij}$ by counting token positions in which both features appear in the top-$K$ active set across the same corpus. If polynomial interactions merely replicated bigram statistics, $B_{ij}$ and $N_{ij}$ would correlate strongly. As a baseline, we consider the empirical covariance of SAE activations, which captures the full pairwise structure accessible to a linear model.

As expected, this covariance correlates strongly with co-occurrence frequency ($r = 0.82$).
In contrast, PolySAE’s learned interactions exhibit negligible correlation with co-occurrence ($r = 0.06$), indicating that interaction capacity is allocated based on criteria largely orthogonal to frequency.

Since higher-order dictionaries do not simply encode co-occurrence, we ask \textit{(Q7) whether the learned interactions are interpretable}.
To analyze this structure, we construct a dictionary mapping each feature to its most activating tokens, then examine feature pairs and triples with high interaction strength by extracting representative contexts in which the corresponding features co-activate.

Selected examples in \Cref{tab:polysae_second_order} and \Cref{tab:polysae_third_order} illustrate the qualitative structure captured by PolySAE’s higher-order terms.
Second-order interactions often correspond to coherent phrase-level compositions that are not recoverable from either feature in isolation, such as \emph{coffee} $\times$ \emph{star} yielding contexts referring to \emph{Starbucks}, a highly non-linear semantic mapping.
In contrast, SAEs typically activate broad or weakly related features in these contexts, failing to recover the composed meaning.
Third-order interactions further refine such compositions by conditioning on additional context.
For example, PolySAE distinguishes financial \emph{investing} from unrelated \texttt{-ing} usages by integrating morphological cues with market-related features, and disambiguates generic entities such as \emph{nuclear} or \emph{Americans} based on surrounding semantic attributes.
Across examples, \textbf{higher-order terms absorb contextual variation that would otherwise fragment linear features}, allowing PolySAE to express compositional meaning through structured interactions rather than proliferating context-specific atoms. Further examples in \Cref{sec:appendix_qualitative} confirm these patterns.

\textbf{Quantifying interpretability at scale.}
The qualitative examples above show that selected interactions are interpretable; we now ask \textit{(Q8) what fraction of PolySAE's learned interactions are interpretable}. We query GPT-4o-mini with a structured prompt that returns a 0--1 score for how strongly each interaction matches the composite concept implied by its two constituent features (\Cref{app:llm_judge}). Of the $\sim\!5 \times 10^7$ candidate pairs over the top 10K features by activation mass, 292{,}361 exhibit non-negligible interaction strength $B_{ij}$. Of the 70,000 pairs we evaluated, 8{,}550 score above 0.9, a 12\% rate of highly interpretable compositional interactions. Even assuming not every one of the 16K linear features is itself interpretable, \textit{PolySAE adds at least 8{,}550 compositional concepts at the second-order level alone on top of the linear dictionary}.

\subsection{Steering with Learned Interactions}
\label{sec:qualitative}

Having established that the learned interactions are interpretable; we now ask \textit{(Q9) whether the directions they define are causally useful for steering model behavior}. For each of 27 compositional concepts drawn from \Cref{tab:polysae_second_order} and \Cref{sec:appendix_qualitative}, we form a steering vector $d_i + d_j$ from the two corresponding decoder directions and inject $\alpha(d_i + d_j)$ into GPT-2's layer-8 residual stream at every token position during autoregressive generation (50 tokens). For each concept we design 12 neutral prompts in which the compositional target could plausibly appear (e.g., \textit{``The controversial cross-border pipeline project is called the''}), giving 324 prompt-concept pairs, and compare PolySAE directions against vanilla SAE directions and no steering. Each condition is evaluated by the rank of the target compositional token in the model's next-token distribution; lower rank is better.

\textbf{Steering with compositional directions.}
PolySAE \textit{shifts generation toward the compositional target far more reliably} than vanilla SAE. Across all 324 prompt-concept pairs, PolySAE achieves lower target rank than no steering in 230 cases (71.0\%) with only 3 degradations. Aggregating to the concept level, PolySAE improves over no steering in 27/27 concepts and over vanilla SAE in 21/27, with a mean rank improvement of $+41.5$ against vanilla. Representative examples appear in \Cref{tab:steering_examples_main}: steering with \texttt{[canada]}$\times$\texttt{[oil]} produces \textit{Keystone XL} under PolySAE, while vanilla produces \textit{Trans Mountain pipeline} and no steering defaults to the unrelated \textit{Trans-Pacific Partnership}. Across concepts, vanilla SAE often fails to shift output away from the unsteered baseline at all, while PolySAE recovers the intended composition.

\textbf{Direction alignment.}
We verify that these gains reflect better-aligned representations. For each concept, we estimate a ground-truth compositional direction using the difference-in-means method \citep{marks2024geometry}, which AxBench \citep{wu2025axbench} demonstrated to be among the most effective approaches for identifying causally efficacious concept directions. We collect 20 sentences containing the concept and 20 without, extract GPT-2 layer-8 activations at the target token, and compute $\hat{d}_{\text{concept}} = \mathrm{normalize}(\bar{a}_{\text{pos}} - \bar{a}_{\text{neg}})$, where $\bar{a}_{\text{pos}}$ and $\bar{a}_{\text{neg}}$ are the mean activations across the positive and negative sets. We then measure cosine similarity between each model's $d_i + d_j$ and this ground truth. PolySAE achieves mean cosine similarity $0.372 \pm 0.093$ vs.\ $0.311 \pm 0.158$ for vanilla SAE, a 19.7\% relative improvement (11/27 wins vs.\ 2/27, 14 ties).

%% file: figures/topk_width.tex
\begin{figure}[!t]
\centering
\begin{tikzpicture}
\begin{axis}[
    width=0.95\columnwidth,
    height=0.65\columnwidth,
    xlabel={Sparsity $k$ (active features)},
    ylabel={Mean Probing F1},
    xmode=log,
    log basis x=2,
    xtick={4,8,16,32,64,128},
    xticklabels={4,8,16,32,64,128},
    ymin=0.55, ymax=0.82,
    ymajorgrids=true,
    yminorgrids=true,
    minor y tick num=2,
    grid style={line width=.1pt, draw=gray!15},
    legend style={font=\small, at={(0.02,0.98)}, anchor=north west},
]
\addplot[fill=blue!15, draw=none, forget plot] coordinates {
    (4, 0.602) (8, 0.620) (16, 0.654) (32, 0.660) (64, 0.641) (128, 0.655) 
    (128, 0.764) (64, 0.686) (32, 0.685) (16, 0.701) (8, 0.662) (4, 0.659)
} -- cycle;
\addplot[thick, color=blue!70, mark=o, mark size=2pt] coordinates {
    (4, 0.632) (8, 0.637) (16, 0.676) (32, 0.673) (64, 0.660) (128, 0.690)
};
\addlegendentry{SAE}
\addplot[fill=red!15, draw=none, forget plot] coordinates {
    (4, 0.634) (8, 0.675) (16, 0.695) (32, 0.711) (64, 0.749) (128, 0.777) 
    (128, 0.801) (64, 0.790) (32, 0.760) (16, 0.715) (8, 0.700) (4, 0.697)
} -- cycle;
\addplot[thick, color=red!70, mark=*, mark size=2pt] coordinates {
    (4, 0.663) (8, 0.692) (16, 0.705) (32, 0.742) (64, 0.776) (128, 0.786)
};
\addlegendentry{PolySAE}
\end{axis}
\end{tikzpicture}
\caption{\textbf{Probing Mean F1 vs.\ sparsity $k$}. Shaded regions show range across widths (2k--16k). 
PolySAE consistently outperforms SAE with significant separation at higher $k$.}
\label{fig:sparsity}
\end{figure}

%% file: sections/6.conclusion.tex
\section{Conclusion}\label{sec:conclusion}
We introduced PolySAE, a sparse autoencoder that extends the decoder with higher-order terms to model feature interactions while preserving a linear encoder for interpretability. Through low-rank tensor factorization on a shared projection subspace, PolySAE captures pairwise and triple interactions with small parameter overhead. Across four LLMs and three SAE variants, it improves probing F1 by 8\% on average while maintaining comparable reconstruction, achieves 2--10$\times$ larger Wasserstein distances between class-conditional distributions, and allocates interaction capacity based on compositional structure rather than co-occurrence ($r=0.06$). The learned interaction directions causally steer model outputs toward the corresponding compositional semantics. \textbf{Limitations:} we study models up to 2B parameters and restrict experiments to forced-sparsity SAE variants.

%% file: sections/appendix.tex
\section{PolySAE Algorithm and Codebase}

We provide the full training algorithm for PolySAE in \Cref{alg:polysae}, detailing the encoding, polynomial decoding, and optimization steps used throughout all experiments. Our codebase is publicly available at  \url{https://github.com/pakoromilas/PolySAE}

\begin{algorithm}[t]
\caption{PolySAE Training}
\label{alg:polysae}
\begin{algorithmic}
  \STATE {\bfseries Input:} activations $\{\vx\}$, ranks $(R_1,R_2,R_3)$, sparsity $K$, learning rate $\eta$, rescale by decoder norm flag $\rho$
  \STATE Initialize $\mU \gets \mathrm{qr}_+(\mU_{\mathrm{rand}})$; \quad $\lambda_2 \gets -0.5$; \quad $\lambda_3 \gets 0.5$
  \FOR{each minibatch $\vx$}
    \STATE {\emph{// Encode}}
    \STATE $\vh \gets \mE^\top\vx + \vb_{\mathrm{enc}}$
    \IF{$\rho$}
      \STATE Compute decoder norms $\vd \in \mathbb{R}^{d_{\text{sae}}}$: $d_i = \|\text{PolyDec}(\ve_i)\|_2$
      \STATE $\vz \gets \mathrm{TopK}\!\left(\mathrm{ReLU}(\vh \odot \vd), K\right)$
      \STATE $\vz \gets \vz / \vd$ \qquad {\emph{// rescale back for decode invariance}}
    \ELSE
      \STATE $\vz \gets \mathrm{TopK}\!\left(\mathrm{ReLU}(\vh), K\right)$
    \ENDIF
    \STATE {\emph{// Decode (polynomial, shared, hierarchical)}}
    \STATE $\vy_1 \gets (\vz\,\mU)\,\mC^{(1)\!\top}$
    \STATE $\mA_2 \gets \vz\,\mU_{:,1:R_2}$
    \STATE $\vy_2 \gets (\mA_2 * \mA_2)\,\mC^{(2)\!\top}$
    \STATE $\mA_3 \gets \vz\,\mU_{:,1:R_3}$
    \STATE $\vy_3 \gets (\mA_3 * \mA_3 * \mA_3)\,\mC^{(3)\!\top}$
    \STATE $\vy \gets \vb_{\mathrm{dec}} + \vy_1 + \lambda_2\,\vy_2 + \lambda_3\,\vy_3$
    \STATE {\emph{// Update with manifold retraction}}
    \STATE $\mathcal{L} \gets \|\vy - \vx\|_2^2 + \text{regularizations}$
    \STATE Update all parameters via $\nabla \mathcal{L}$
    \STATE $(\mQ, \mR) \gets \mathrm{qr}(\mU)$; \quad $\mS \gets \mathrm{diag}(\mathrm{sgn}(\mathrm{diag}(\mR)))$; \quad $\mU \gets \mQ\mS$
  \ENDFOR
\end{algorithmic}
\end{algorithm}

\section{Implementation Details}
\label{sec:appendix_implementation}

\paragraph{Architecture and Sparsification.}
We train standard sparse autoencoders (SAEs) and PolySAEs with a latent width of $16{,}384$ and sparsity level $K=64$.
Encoders use one of three sparsification strategies: Top-$K$ \citep{gao2025scaling}, BatchTopK \citep{bussmann2024batchtopk}, or Matryoshka \citep{bussmann2025matryoshka}.
For Top-$K$ and BatchTopK, the $K$ largest activations per token (or batch) are retained and the remainder zeroed.
All models are trained on residual-stream activations extracted from pretrained language models.

\paragraph{LLMs.}
We evaluate SAEs and PolySAEs on a standard set of pretrained language models spanning a range of scales: GPT-2 Small \citep{radford2019gpt2}, Pythia-410M and Pythia-1.4B \citep{biderman2023pythia}, and Gemma-2-2B \citep{gemma2report}.
For each model, we extract residual-stream activations from a single transformer layer chosen near the center of the network, following the methodology of \citet{dunefsky2024transcoders}.

\paragraph{PolySAE Decoder Ranks.}
The rank configurations used in our experiments are:
\begin{itemize}[leftmargin=*, itemsep=2pt]
    \item GPT-2 Small \citep{radford2019gpt2}: $(768, 64, 64)$
    \item Pythia-410M \citep{biderman2023pythia}: $(1024, 128, 128)$
    \item Pythia-1.4B \citep{biderman2023pythia}: $(2048, 128, 128)$
    \item Gemma-2-2B \citep{gemma2report}: $(2304, 128, 128)$
\end{itemize}

\paragraph{Training Setup.} All models are trained using the Adam optimizer with $\beta_1 = 0.9$ and $\beta_2 = 0.999$, a constant learning rate of $3\times10^{-4}$ with no warmup or decay schedules. We use a batch size of $4096$ tokens and a context length of $128$. We apply gradient clipping with a maximum norm of $1.0$ to stabilize training. No weight decay or L1 regularization is applied to the encoder or decoder weights. Training runs for $5\times10^8$ tokens for Gemma-2-2B and Pythia models, and $3\times10^8$ tokens for GPT-2 Small, following the protocol used in the main experiments. 

\paragraph{Datasets.}
For Gemma-2-2B and GPT-2 Small, training data is drawn from OpenWebText \citep{gokaslan2019openwebtext}.
For Pythia-410M and Pythia-1.4B, we use an uncopyrighted variant of the deduplicated Pile \citep{gao2021pile}.
Reconstruction is evaluated on held-out data from the same distribution as training.

\paragraph{Evaluation.}
We evaluate learned representations using SAEBench \citep{karvonen2025saebench}.
Reported metrics include reconstruction error on held-out data and sparse probing performance on six classification tasks:
Bias in Bios \citep{dearteaga2019biasinbios},
AG News \citep{zhang2015charcnn},
EuroParl \citep{koehn-2005-europarl},
GitHub programming languages \citep{codeparrot_github_code},
Amazon Sentiment, and Amazon-15 \citep{hou2024bridging}.

\paragraph{Implementation.}
Our training pipeline extends \texttt{SAELens} \citep{bloom2024saetrainingcodebase} to support PolySAE while preserving the standard SAE training interface.
PolySAE differs from standard SAEs only in the decoder; all other components, including the encoder, sparsification strategy, optimizer, and evaluation pipeline, are shared across models.

\section{Extended Qualitative Analysis}
\label{sec:appendix_qualitative}

We present an extended qualitative analysis of the interaction structure learned by PolySAE.
The analysis proceeds hierarchically, first examining second-order (pairwise) interactions and then extending to third-order (triplet) compositions.
Throughout, we compare PolySAE to a vanilla Top-$K$ SAE trained under identical conditions.

\subsection{Second-Order Analysis}

We begin by analyzing pairwise interactions to assess whether PolySAE captures compositional structure beyond surface-level co-occurrence.

\paragraph{Setup.}
Both models are applied to 1M documents from OpenWebText.
Features are ranked by total activation mass, and the top 10{,}000 are retained, yielding approximately $5\times10^{7}$ candidate feature pairs.

\paragraph{Interaction Strength.}
For PolySAE, we quantify the strength of a feature pair $(i,j)$ using the learned quadratic decoder weights:
\begin{equation}
B_{ij} = \lambda_2 \left\| (\vu_i \odot \vu_j)^\top \mC^{(2)} \right\|_2 ,
\end{equation}
which reflects how much decoder capacity is assigned to that interaction.
For the vanilla SAE, which lacks explicit interaction parameters, we use empirical feature covariance,
\begin{equation}
\mathrm{Cov}_{ij} = \mathbb{E}[z_i z_j] - \mathbb{E}[z_i]\mathbb{E}[z_j],
\end{equation}
as a proxy for pairwise structure.

\paragraph{Relation to Co-occurrence.}
We independently estimate empirical co-occurrence by counting positions where both features appear in the Top-$K$ active set.
For the vanilla SAE, covariance is strongly correlated with co-occurrence ($r=0.82$), indicating that pairwise structure largely mirrors frequency.
In contrast, PolySAE’s interaction strengths show negligible correlation with co-occurrence ($r=0.06$), suggesting that learned interactions reflect structure beyond surface statistics.

\paragraph{Qualitative Regimes.}
The weak coupling between interaction strength and frequency allows us to identify qualitatively distinct regimes.
Of particular interest are \emph{latent} interactions: feature pairs with strong learned interactions despite low empirical co-occurrence.
These pairs often correspond to meaningful compositional patterns that are not recoverable from frequency alone.

\paragraph{Examples.}
For interactions above the 80th percentile in $B_{ij}$, we extract representative contexts in which both features co-activate.
We mark the target token in each sentence and label features by their top-activating tokens.
Comparing these contexts with vanilla SAE activations highlights cases where PolySAE captures relationships that the linear model does not.

\subsection{Third-Order Analysis}

We next examine whether third-order interactions refine or disambiguate pairwise compositions.

\paragraph{Candidate Selection.}
We focus on latent second-order pairs—those with high interaction strength but low co-occurrence—and identify corpus positions where both features are simultaneously active.

\paragraph{Triplet Scoring.}
Within these contexts, we evaluate all co-active third features using the learned cubic decoder:
\begin{equation}
\mathrm{Gamma}(f_1, f_2, k)
= \lambda_3 \left| (\vu_{f_1} \odot \vu_{f_2}) \cdot \vu_k^\top \mC^{(3)\top} \right| .
\end{equation}
For each pair, we retain the third feature with the highest score, after filtering stopword-like features.

\paragraph{Interpretation.}
The resulting triplets are consistently interpretable, with the third feature modulating the meaning of the pair rather than introducing unrelated content.
Common patterns include entity--attribute--domain and subject--object--context structures.
Representative examples are shown in \cref{tab:polysae_third_order_main}, illustrating how higher-order interactions sharpen and contextualize pairwise compositions.

\subsection{Additional Second-Order Interaction Examples}

Tables~\ref{tab:polysae_main}--\ref{tab:polysae_appendix_compounds} show additional second-order interaction examples.
Each row highlights a token where two PolySAE features are simultaneously active.
Across these tables, the interacting features are typically more specific than the corresponding vanilla SAE features in the same context.
The vanilla SAE often activates on a single high-level, morphological, or broadly related feature, while PolySAE activations reflect a more refined decomposition at the highlighted token.

\subsection{Additional Third-Order Interaction Examples}
\label{app:third_order_interactions}

Tables~\ref{tab:polysae_third_order_main} and \ref{tab:polysae_third_order_appendix} present further third-order examples.
Each row shows contexts in which three PolySAE features co-activate at the same token.
In these cases, the activated features vary with context and appear more specific than the corresponding vanilla SAE activations, which often capture only one component or default to generic features.


\begin{table*}[t]
\centering
\small
\setlength{\tabcolsep}{4pt}
\renewcommand{\arraystretch}{1.05}
\caption{\textbf{Compositional Interactions Captured by PolySAE.} PolySAE features (A and B) bind in context to represent specific compositional concepts. Vanilla SAE features (with high-frequency features filtered) fail to capture these compositions.}
\label{tab:polysae_main}

\begingroup
\setlength{\emergencystretch}{2em}
\sloppy

\begin{tabularx}{\textwidth}{@{}L{2.6cm} L{2.6cm} Y L{2.6cm}@{}}
\toprule
\textbf{Poly Feature A} &
\textbf{Poly Feature B} &
\textbf{Context } &
\textbf{Vanilla SAE} \\
\midrule

[star,\hspace{0pt} stars] &
[coffee,\hspace{0pt} tea] &
We've all certainly heard of beers brewed with espresso, but how about one with an espresso shot poured over the top? \textbf{Starbucks} &
[Apple,\hspace{0pt} Google] \\
\addlinespace

[officially,\hspace{0pt} ically] &
[traditional,\hspace{0pt} conventional] &
It's hard to say what's most impressive about Eduardo Garcia. The class\textbf{ically}-trained chef spent years cooking aboard yachts &
[newly,\hspace{0pt} ly] \\
\addlinespace

[surgery,\hspace{0pt} repair] &
[Trans,\hspace{0pt} LGBT] &
Some in the transgender community are worried a suspicious fire at a Montreal clinic will add delays to an already lengthy process to get gender reassignment \textbf{surgery} &
[birth,\hspace{0pt} baby] \\
\addlinespace

[DNA,\hspace{0pt} genetic] &
[mod,\hspace{0pt} mods] &
Activists are opening up a new front in their campaign against genetic modification. The latest target is genetically-\textbf{mod}ified trees, which scientists believe could bring huge sustainability &
[modified,\hspace{0pt} edit] \\
\addlinespace

[secret,\hspace{0pt} hidden] &
[Snowden,\hspace{0pt} WikiLeaks] &
On May 24th PBS aired a Frontline documentary about alleged Wikileaker Bradley Manning called ``\textbf{WikiSecrets}'' &
[secret,\hspace{0pt} secrets] \\
\addlinespace

[business,\hspace{0pt} businesses] &
[man,\hspace{0pt} woman] &
By Joseph George. The business\textbf{man} dad of the boy who drove a Ferrari and was arrested by police in Kerala, India &
[man,\hspace{0pt} President] \\

\bottomrule
\end{tabularx}

\endgroup
\end{table*}

\begin{table*}[t]
\centering
\small
\setlength{\tabcolsep}{4pt}
\renewcommand{\arraystretch}{1.05}
\caption{\textbf{PolySAE Interactions -- Brand \& Proper Noun Decomposition.} PolySAE decomposes compound names into their semantic constituents. Vanilla SAE (with high-frequency features filtered) often fires on unrelated entities or only captures the surface form.}
\label{tab:polysae_appendix_proper_nouns}

\begingroup
\setlength{\emergencystretch}{2em}
\sloppy

\begin{tabularx}{\textwidth}{@{}L{2.6cm} L{2.6cm} Y L{2.6cm}@{}}
\toprule
\textbf{Poly Feature A} &
\textbf{Poly Feature B} &
\textbf{Context} &
\textbf{Vanilla SAE} \\
\midrule

[economic,\hspace{0pt} economy] &
[Times,\hspace{0pt} magazine] &
The RBI on Wednesday did not allow Stanley Pignal, the South Asian business and finance correspondent for the \textbf{Economist} magazine, to attend the central bank's &
[economics,\hspace{0pt} economist] \\
\addlinespace

[field,\hspace{0pt} fields] &
[University,\hspace{0pt} school] &
John Doe is a Jesuit with ADHD. He was an outstanding student and a compassionate senior at Fair\textbf{field} University who played sports and volunteered often at a literacy &
[York,\hspace{0pt} Washington] \\
\addlinespace

[Star,\hspace{0pt} Chronicle] &
[staff,\hspace{0pt} crew] &
Man Arrested after Stolen Mower Runs Out of Gas. By West Kentucky Star \textbf{Staff}. PADUCAH, KY &
[staff,\hspace{0pt} faculty] \\
\addlinespace

[Dragon,\hspace{0pt} Iron] &
[steel,\hspace{0pt} Pittsburgh] &
The \textbf{Iron} Horde is on the march, and the Warlords of Draenor are primed to invade Azeroth on November 13! Steel yourself for the onslaught by watching &
[assault,\hspace{0pt} steel] \\
\addlinespace

[Italian,\hspace{0pt} Italy] &
[gang,\hspace{0pt} mob] &
Details obtained by the Guardian reveal extent to which \textbf{Sicilian} mafia clans are migrating north after running into financial problems in Italy. &
[State,\hspace{0pt} ISIS] \\

\bottomrule
\end{tabularx}

\endgroup
\end{table*}

\begin{table*}[t]
\centering
\small
\setlength{\tabcolsep}{4pt}
\renewcommand{\arraystretch}{1.05}
\caption{\textbf{PolySAE Interactions -- Morphological Composition.} PolySAE binds suffix/prefix features with semantic content to form derived words. Vanilla SAE (with high-frequency features filtered) captures only generic morphological patterns without semantic binding.}
\label{tab:polysae_appendix_morphology}

\begingroup
\setlength{\emergencystretch}{2em}
\sloppy

\begin{tabularx}{\textwidth}{@{}L{2.6cm} L{2.6cm} Y L{2.6cm}@{}}
\toprule
\textbf{Poly Feature A} &
\textbf{Poly Feature B} &
\textbf{Context} &
\textbf{Vanilla SAE} \\
\midrule

[ers,\hspace{0pt} Workers] &
[administration,\hspace{0pt} administrative] &
Piedmont High School. A school reveals it has a ``Fantasy Slut League'' \textbf{Administrators} try to do the right thing, but fall woefully short of &
[members,\hspace{0pt} ers] \\
\addlinespace

[ing,\hspace{0pt} ings] &
[arrested,\hspace{0pt} arrest] &
Earlier this year, The Heritage Foundation's Meese Center released Arrest\textbf{ing} Your Property, a comprehensive report on civil asset forfeiture-the much mal &
[ing,\hspace{0pt} training] \\
\addlinespace

[protest,\hspace{0pt} protests] &
[making,\hspace{0pt} ing] &
Major League Baseball can no longer claim to be free of any anthem-protest\textbf{ing} players. On Saturday night, A's catcher Bruce Maxwell took &
[ing,\hspace{0pt} training] \\
\addlinespace

[ers,\hspace{0pt} Workers] &
[photos,\hspace{0pt} pictures] &
In-Sight Film. The film in-sight was produced in conjunction with the Format Photography Festival to mark 10 years of the Street Photograph\textbf{ers} group &
[members,\hspace{0pt} ers] \\
\addlinespace

[ized,\hspace{0pt} ization] &
[treatment,\hspace{0pt} drugs] &
The flu shot is a quack science medical hoax. While some vaccines do confer immuniz\textbf{ation} effectiveness, the flu shot isn't one of them &
[development,\hspace{0pt} ation] \\
\addlinespace

[bound,\hspace{0pt} ice] &
[gun,\hspace{0pt} guns] &
A new Texas law gives gun owners a new right to store a weapon (any lawfully owned firearm, not just those owned under a Concealed Handgun L\textbf{ice}nse &
[gun,\hspace{0pt} weapons] \\

\bottomrule
\end{tabularx}

\endgroup
\end{table*}

\begin{table*}[t]
\centering
\small
\setlength{\tabcolsep}{4pt}
\renewcommand{\arraystretch}{1.05}
\caption{\textbf{PolySAE Interactions -- Domain-Specific Collocations.} PolySAE captures specialized terminology through the interaction of domain features. Vanilla SAE (with high-frequency features filtered) often misses the domain-specific meaning.}
\label{tab:polysae_appendix_domain}

\begingroup
\setlength{\emergencystretch}{2em}
\sloppy

\begin{tabularx}{\textwidth}{@{}L{2.6cm} L{2.6cm} Y L{2.6cm}@{}}
\toprule
\textbf{Poly Feature A} &
\textbf{Poly Feature B} &
\textbf{Context} &
\textbf{Vanilla SAE} \\
\midrule

[football,\hspace{0pt} NFL] &
[conference,\hspace{0pt} conferences] &
Statement from the Southeastern Conference Office Regarding the Florida-LSU football game: The LSU-Florida \textbf{football} game scheduled for Saturday in Gainesville &
[League,\hspace{0pt} Conference] \\
\addlinespace

[earnings,\hspace{0pt} financial] &
[number,\hspace{0pt} numbers] &
T-Mobile US, Inc.\ TMUS is scheduled to report fourth-quarter 2015 financial \textbf{number}s, before the opening bell on Feb 17. Last &
[numbers,\hspace{0pt} figures] \\
\addlinespace

[technology,\hspace{0pt} tech] &
[development,\hspace{0pt} developers] &
You can't look at internet news lately without seeing the latest and greatest in nanotechnology \textbf{development}s. Everything these days is being manufactured smaller, faster &
[it,\hspace{0pt} said] \\
\addlinespace

[Canada,\hspace{0pt} Canadian] &
[oil,\hspace{0pt} pipeline] &
Eddy Radillo holds a Texas flag and a sign opposing the Transcanada Keystone \textbf{Pipeline} in February 2012 outside the Lamar County Courthouse in Paris &
[oil,\hspace{0pt} gas] \\
\addlinespace

[diet,\hspace{0pt} fitness] &
[train,\hspace{0pt} rail] &
Here's what you need to know... Your gains will stagnate if you only weight \textbf{train} within the same rep ranges and loading patterns. &
[training,\hspace{0pt} train] \\

\bottomrule
\end{tabularx}

\endgroup
\end{table*}

\begin{table*}[t]
\centering
\small
\setlength{\tabcolsep}{4pt}
\renewcommand{\arraystretch}{1.05}
\caption{\textbf{PolySAE Interactions -- Compound Words \& Phrases.} PolySAE captures compound words and multi-word phrases through feature interactions. Vanilla SAE (with high-frequency features filtered) often misses the compositional meaning entirely.}
\label{tab:polysae_appendix_compounds}

\begingroup
\setlength{\emergencystretch}{2em}
\sloppy

\begin{tabularx}{\textwidth}{@{}L{2.6cm} L{2.6cm} Y L{2.6cm}@{}}
\toprule
\textbf{Poly Feature A} &
\textbf{Poly Feature B} &
\textbf{Context} &
\textbf{Vanilla SAE} \\
\midrule

[director,\hspace{0pt} founder] &
[lead,\hspace{0pt} managing] &
A FRIEND OF MINE recently made the following observation about Ezra Koenig, the founder and \textbf{lead} singer of Vampire Weekend. ``Did you realize, &
[led,\hspace{0pt} lead] \\
\addlinespace

[written,\hspace{0pt} designed] &
[research,\hspace{0pt} researcher] &
The following story was written and \textbf{researched} by Rone Tempest for The Utah Investigative Journalism Project in partnership with The Salt Lake Tribune. Dustin Porter said &
[created,\hspace{0pt} made] \\
\addlinespace

[alleged,\hspace{0pt} allegations] &
[level,\hspace{0pt} levels] &
Back to previous page. Accusations against generals cast dark shadow over Army. By Ernesto Londo\~no. The accusations \textbf{leve}led against &
[place,\hspace{0pt} made] \\
\addlinespace

[music,\hspace{0pt} musical] &
[official,\hspace{0pt} officer] &
ROCHESTER, N.Y.\ -- Members of Rochester's music community continue to pull together to remember and help the family a fellow \textbf{musician} who &
[man,\hspace{0pt} President] \\
\addlinespace

[involved,\hspace{0pt} involvement] &
[support,\hspace{0pt} help] &
STEAL THIS SHOW's Patreon campaign helps keep us free and independent. If you enjoy the show, get \textbf{involve}d. Our patrons get access to &
[started,\hspace{0pt} ready] \\
\addlinespace

[shooting,\hspace{0pt} shot] &
[focus,\hspace{0pt} focused] &
Berenice Abbott was an American photographer best known for her black-and-white photography of New York City. She heavily focused her \textbf{shooting} &
[ing,\hspace{0pt} training] \\
\addlinespace

[document,\hspace{0pt} documents] &
[content,\hspace{0pt} contents] &
Use these links to rapidly review the \textbf{document}. TABLE OF CONTENTS. INDEX TO CONSOLIDATED FINANCIAL STATEMENTS &
[Introduction,\hspace{0pt} History] \\

\bottomrule
\end{tabularx}

\endgroup
\end{table*}

\begin{table*}[t]
\centering
\small
\setlength{\tabcolsep}{3pt}
\renewcommand{\arraystretch}{1.05}
\caption{\textbf{Third-Order Compositional Interactions Captured by PolySAE.} Three PolySAE features ($F_i$, $F_j$, $F_k$) bind in context to represent compositional concepts. Vanilla SAE often captures individual components but misses the compositional structure.}
\label{tab:polysae_third_order_main}

\begingroup
\setlength{\emergencystretch}{2em}
\sloppy

\begin{tabularx}{\textwidth}{@{}L{2.0cm} L{2.0cm} L{2.0cm} Y L{2.2cm}@{}}
\toprule
\textbf{Poly $F_i$} &
\textbf{Poly $F_j$} &
\textbf{Poly $F_k$} &
\textbf{Context} &
\textbf{Vanilla SAE} \\
\midrule

[nuclear,\hspace{0pt} Fukushima,\hspace{0pt} reactor] &
[test,\hspace{0pt} testing,\hspace{0pt} tested] &
[radiation,\hspace{0pt} laser,\hspace{0pt} magnetic] &
US tests \textbf{nuclear}-capable missile with the range to strike North Korea. The US has test-fired a nuclear-capable intercontinental ballistic missile &
[nuclear,\hspace{0pt} reactor,\hspace{0pt} atomic] \\
\addlinespace

[black,\hspace{0pt} white,\hspace{0pt} racial] &
[Americans,\hspace{0pt} Canadians,\hspace{0pt} Australians] &
[people,\hspace{0pt} women,\hspace{0pt} men] &
In a push to get more Black \textbf{Americans} involved in the world of tech, a slew of organizations have teamed up with South by Southwest &
[Americans,\hspace{0pt} Muslims,\hspace{0pt} Jews] \\
\addlinespace

[ing,\hspace{0pt} ings,\hspace{0pt} ting] &
[stock,\hspace{0pt} trading,\hspace{0pt} market] &
[investment,\hspace{0pt} invest,\hspace{0pt} investing] &
Philippines stocks higher at close of trade; PSEi Composite up 0.57\%. \textbf{Invest}ing.com --- Philippines stocks were higher after &
[ing,\hspace{0pt} training,\hspace{0pt} running] \\
\addlinespace

[line,\hspace{0pt} lines,\hspace{0pt} lining] &
[supply,\hspace{0pt} supplies,\hspace{0pt} shortage] &
[road,\hspace{0pt} route,\hspace{0pt} pipeline] &
The same is true of supply \textbf{lines} into landlocked Afghanistan. Within months of the 2001 invasion, Mr.\ Musharraf signed a deal &
[the,\hspace{0pt} ,,\hspace{0pt} .,\hspace{0pt} ',\hspace{0pt} of]\newline [the,\hspace{0pt} ,,\hspace{0pt} ',\hspace{0pt} of,\hspace{0pt} a] \\
\addlinespace

[proved,\hspace{0pt} proven,\hspace{0pt} prove] &
[star,\hspace{0pt} stars,\hspace{0pt} superstar] &
[reputation,\hspace{0pt} popularity,\hspace{0pt} fame] &
Arguably the biggest surprise would have been if he had turned up, but David Bowie proved some \textbf{stars} are big enough not to have make themselves available &
[star,\hspace{0pt} stars,\hspace{0pt} superstar] \\
\addlinespace

[historic,\hspace{0pt} historical,\hspace{0pt} historically] &
[UFC,\hspace{0pt} fight,\hspace{0pt} MMA] &
[strong,\hspace{0pt} impressive,\hspace{0pt} solid] &
After 1,501 days as UFC light-heavyweight champion, Jon Jones' \textbf{historic} title reign came to an end late Tuesday when he was stripped &
[the,\hspace{0pt} ,,\hspace{0pt} .,\hspace{0pt} ',\hspace{0pt} of] \\
\addlinespace

[treated,\hspace{0pt} treat,\hspace{0pt} treating] &
[consumers,\hspace{0pt} consumer,\hspace{0pt} consumption] &
[customers,\hspace{0pt} customer,\hspace{0pt} clients] &
Jeremy Corbyn today warned the banking industry it must not treat \textbf{consumers} and entrepreneurs as a ``cash cow'' and attacked the links between senior politicians &
[the,\hspace{0pt} ,,\hspace{0pt} .,\hspace{0pt} ',\hspace{0pt} of] \\

\bottomrule
\end{tabularx}

\endgroup
\end{table*}


\begin{table*}[t]
\centering
\small
\setlength{\tabcolsep}{3pt}
\renewcommand{\arraystretch}{1.05}
\caption{\textbf{Additional Third-Order PolySAE Interactions.} Further examples of three-way feature compositions. Vanilla SAE sometimes captures individual components but misses the compositional structure.}
\label{tab:polysae_third_order_appendix}

\begingroup
\setlength{\emergencystretch}{2em}
\sloppy

\begin{tabularx}{\textwidth}{@{}L{2.0cm} L{2.0cm} L{2.0cm} Y L{2.2cm}@{}}
\toprule
\textbf{Poly $F_i$} &
\textbf{Poly $F_j$} &
\textbf{Poly $F_k$} &
\textbf{Context} &
\textbf{Vanilla SAE} \\
\midrule

[Army,\hspace{0pt} Force,\hspace{0pt} Navy] &
[Israel,\hspace{0pt} Israeli,\hspace{0pt} Jewish] &
[IDF] &
Earlier this week, the Friends of the Israel Defense Forces, an organization dedicated to supporting the men and women serving in the \textbf{IDF}, held its annual dinner &
[Israel,\hspace{0pt} Israeli,\hspace{0pt} Jewish]\newline [the,\hspace{0pt} ,,\hspace{0pt} .,\hspace{0pt} ',\hspace{0pt} of] \\
\addlinespace

[annual,\hspace{0pt} monthly,\hspace{0pt} annually] &
[percent,\hspace{0pt} \%,\hspace{0pt} points] &
[regular,\hspace{0pt} regularly,\hspace{0pt} frequent] &
According to the latest research from our Wireless Smartphone Strategies (WSS) service, global smartphone shipments grew 6 percent \textbf{annually} to reach 360 million units &
[annual,\hspace{0pt} monthly,\hspace{0pt} annually]\newline [the,\hspace{0pt} ,,\hspace{0pt} .,\hspace{0pt} ',\hspace{0pt} of] \\
\addlinespace

[get,\hspace{0pt} make,\hspace{0pt} getting] &
[film,\hspace{0pt} movie,\hspace{0pt} films] &
[documented,\hspace{0pt} depicted,\hspace{0pt} depicts] &
Three tips on how to \textbf{film} anywhere; slums, red light districts, museums, exhibitions, churches, and not get your video camera gear stolen &
[film,\hspace{0pt} movie,\hspace{0pt} films]\newline [the,\hspace{0pt} ,,\hspace{0pt} .,\hspace{0pt} ',\hspace{0pt} of] \\
\addlinespace

[well,\hspace{0pt} ill,\hspace{0pt} poorly] &
[widely,\hspace{0pt} commonly,\hspace{0pt} widespread] &
[best,\hspace{0pt} better,\hspace{0pt} good] &
April 6, 2014. CR Sunday Interview: Zack Soto. ***** is a widely \textbf{well}-liked cartoonist, publisher and &
[well,\hspace{0pt} poorly,\hspace{0pt} badly]\newline [the,\hspace{0pt} ,,\hspace{0pt} .,\hspace{0pt} ',\hspace{0pt} of] \\
\addlinespace

[accept,\hspace{0pt} accepted,\hspace{0pt} accepting] &
[final,\hspace{0pt} ultimate,\hspace{0pt} preliminary] &
[great,\hspace{0pt} considerable,\hspace{0pt} significant] &
NEW YORK --- Dedicated Hillary Clinton supporters accepted \textbf{final} defeat Wednesday morning even as they struggled to accept that their candidate lost &
[final,\hspace{0pt} finals,\hspace{0pt} ultimate]\newline [the,\hspace{0pt} ,,\hspace{0pt} .,\hspace{0pt} ',\hspace{0pt} of] \\
\addlinespace

[percent,\hspace{0pt} \%,\hspace{0pt} percentage] &
[currency,\hspace{0pt} dollar,\hspace{0pt} euro] &
[cents] &
The Canadian dollar dipped below 75 \textbf{cents} (U.S.) in Tuesday's trading as equity markets worldwide remained extremely volatile &
[the,\hspace{0pt} ,,\hspace{0pt} .,\hspace{0pt} ',\hspace{0pt} of]\newline [.,\hspace{0pt} \$,\hspace{0pt} ,,\hspace{0pt} on,\hspace{0pt} to] \\
\addlinespace

[identified,\hspace{0pt} identify,\hspace{0pt} diagnosed] &
[virus,\hspace{0pt} Ebola,\hspace{0pt} HIV] &
[label,\hspace{0pt} labels,\hspace{0pt} labeled] &
the governor of New York State announced that the first case of Ebola had been \textbf{diagnosed} at Bellevue &
[the,\hspace{0pt} ,,\hspace{0pt} .,\hspace{0pt} ',\hspace{0pt} of] \\
\addlinespace

[base,\hspace{0pt} bases,\hspace{0pt} baseline] &
[fans,\hspace{0pt} fan,\hspace{0pt} supporters] &
[demand,\hspace{0pt} turnout,\hspace{0pt} attendance] &
It's a shared problem among fan \textbf{bases} across the National Hockey League: They watch their own players so closely that, after a while &
[the,\hspace{0pt} ,,\hspace{0pt} .,\hspace{0pt} ',\hspace{0pt} of]\newline [the,\hspace{0pt} ,,\hspace{0pt} ',\hspace{0pt} of,\hspace{0pt} a] \\
\addlinespace

[unique,\hspace{0pt} distinct,\hspace{0pt} distinctive] &
[two,\hspace{0pt} different,\hspace{0pt} three] &
[separate,\hspace{0pt} separated,\hspace{0pt} distinction] &
For their collaborative project Jus Now, U.K.\ producer Sam Interface and Trinidad producer LAZAbeam find singularity in mashing up two \textbf{distinct} &
[people,\hspace{0pt} men,\hspace{0pt} officers]\newline [the,\hspace{0pt} ,,\hspace{0pt} .,\hspace{0pt} ',\hspace{0pt} of] \\
\addlinespace

[line,\hspace{0pt} lines,\hspace{0pt} lining] &
[supply,\hspace{0pt} supplies,\hspace{0pt} shortage] &
[road,\hspace{0pt} route,\hspace{0pt} pipeline] &
The same is true of supply \textbf{lines} into landlocked Afghanistan. Within months of the 2001 invasion, Mr.\ Musharraf signed a deal &
[the,\hspace{0pt} ,,\hspace{0pt} .,\hspace{0pt} ',\hspace{0pt} of]\newline [the,\hspace{0pt} ,,\hspace{0pt} ',\hspace{0pt} of,\hspace{0pt} a] \\
\addlinespace

[largest,\hspace{0pt} most,\hspace{0pt} biggest] &
[able,\hspace{0pt} ible,\hspace{0pt} ability] &
[stable,\hspace{0pt} stability,\hspace{0pt} flexible] &
Groundwater, the globe's most depend\textbf{able} water insurance system, is not as renewable as researchers once thought &
[the,\hspace{0pt} ,,\hspace{0pt} .,\hspace{0pt} ',\hspace{0pt} of]\newline [the,\hspace{0pt} ,,\hspace{0pt} ',\hspace{0pt} of,\hspace{0pt} a]\\

\bottomrule
\end{tabularx}

\endgroup
\end{table*}

\section{LLM-as-Judge Prompt for Interaction Interpretability}
\label{app:llm_judge}

To quantify the interpretability of PolySAE's learned interactions at scale (\Cref{sec:qualitative}, Q7), we query GPT-4o-mini with a structured prompt that returns a 0--1 score for how strongly each interaction pair matches the composite concept implied by its two constituent features. For each feature pair $(i,j)$ with non-negligible interaction strength $B_{ij}$, we supply the top-activating tokens of features $i$ and $j$ together with representative corpus contexts in which both features co-activate. The model rates each context independently and returns scores in JSON format. Of the 292{,}361 pairs with non-negligible $B_{ij}$, we evaluated 70K, of which 8{,}550 (12\%) scored above 0.9. The full prompt is given in \Cref{fig:llm_judge_prompt}.

\begin{figure}[t]
\centering
\begin{minipage}{0.95\columnwidth}
\small
\ttfamily
Give each example sentence a 0 to 1 score for how strongly it matches the composite concept implied by features i and j.\\[3pt]
Feature i representative tokens: \{token\_examples\_i\}\\
Feature j representative tokens: \{token\_examples\_j\}\\
Example sentences: \{example\_sentences\}\\[3pt]
Instructions:\\
- Judge how strongly the intended criterion in the question above is satisfied for each example sentence.\\
- 0 means not present at all.\\
- 1 means very strongly present.\\
- If there are multiple example sentences, score each one separately.\\
- Return valid JSON only.\\
- Do not return any explanation or text outside the JSON.\\[3pt]
Return format:\\
\{\{\\
\hspace*{1em}"scores": [0.0]\\
\}\}
\end{minipage}
\caption{Full LLM-as-judge prompt used in \Cref{sec:qualitative} (Q7) to score the interpretability of PolySAE feature interactions.}
\label{fig:llm_judge_prompt}
\end{figure}

\section{Activation Steering: Additional Examples}
\label{app:steering}

We provide additional qualitative examples of activation steering on GPT-2 Small, extending the experiment described in \Cref{sec:qualitative}. \Cref{tab:steering_examples_full} gives the full set of selected examples across the 27 compositional concepts (the main paper shows the first four rows as \Cref{tab:steering_examples_main}). \Cref{tab:steering_polyonly} isolates cases where vanilla SAE directions fail to shift the output from the unsteered baseline at all, while PolySAE directions produce a meaningful semantic change, illustrating the gap captured by the aggregate $+41.5$ mean rank improvement in \Cref{sec:qualitative}.

\begin{table*}[t]
\centering
\caption{Selected examples on activation steering on GPT-2 Small. We add the first-order decoder directions $\mathbf{d}_i + \mathbf{d}_j$ to the residual stream (layer~8) during greedy generation (27 concepts $\times$ 12 prompts = 324 evaluations per condition). Each row shows the continuation under three conditions: No Steering, SAE, and PolySAE. \textbf{Bold} marks where PolySAE steers generation toward the compositional target.}
\label{tab:steering_examples_full}
\small
\setlength{\tabcolsep}{4pt}
\begin{tabular}{p{2.2cm} p{2.4cm} p{3.4cm} p{3.4cm} p{3.4cm}}
\toprule
\textbf{Features $\to$ Target} & \textbf{Prompt} & \textbf{No Steering} & \textbf{SAE} & \textbf{PolySAE} \\
\midrule

\scriptsize\texttt{[surgery]}$\times$\texttt{[trans]} \newline $\to$ \textit{gender}
& ``The procedure that helps individuals align their body with their identity is''
& called ``\textit{body alignment}.'' The procedure involves the body aligning$\ldots$
& called ``\textit{body alignment}.'' The procedure involves the use of a combination$\ldots$
& called ``\textbf{\textit{gender identity surgery}}.'' Performed by a surgeon who specializes in gender$\ldots$ \\
\midrule

\scriptsize\texttt{[canada]}$\times$\texttt{[oil]} \newline $\to$ \textit{Keystone}
& ``The controversial cross-border pipeline project is called the''
& \textit{Trans-Pacific Partnership} (TPP), a major step forward for the U.S.\ and Canada.
& \textit{Trans Mountain pipeline}, a controversial project in the works for years.
& \textbf{\textit{Keystone XL}}. The pipeline would carry crude oil from Alberta to U.S.\ refineries. \\
\midrule

\scriptsize\texttt{[involved]}$\times$\texttt{[support]} \newline $\to$ \textit{community}
& ``The foundation wants more people to become''
& \textit{entrepreneurs}, and it wants to make sure they're not just part of the problem.
& \textit{entrepreneurs}, and it wants to make sure they're not just part of the problem.
& \textbf{\textit{involved in the community}}, and to help them make a difference. \\
\midrule

\scriptsize\texttt{[economic]}$\times$\texttt{[times]} \newline $\to$ \textit{Economist}
& ``The magazine with coverage of world politics and business is The''
& \textit{New York Times}. The magazine with coverage of world politics and business is The New York Times.
& \textit{New York Times}. The New York Times is a daily newspaper in the United States$\ldots$
& \textbf{\textit{Economist}}. The Economist is a magazine that is a global news magazine$\ldots$ \\
\midrule

\scriptsize\texttt{[alleged]}$\times$\texttt{[level]} \newline $\to$ \textit{leveled}
& ``During the hearing criticism was sharply''
& \textit{raised} by the government, which said it was not aware of any evidence$\ldots$
& \textit{raised} by the government of India, accused of being ``bigoted''$\ldots$
& \textbf{\textit{leveled} at the government} for not doing enough to tackle the problem$\ldots$ \\
\midrule

\scriptsize\texttt{[shooting]}$\times$\texttt{[focus]} \newline $\to$ \textit{product}
& ``The model arrived early for the scheduled''
& launch of the new \textit{iPhone 6 and 6 Plus}, delayed until late March.
& launch of the new \textit{iPhone 6 and 6 Plus}, delayed until late March.
& launch of the new \textbf{\textit{Galaxy S8}}, delayed until late March. The Galaxy S8 will be the first$\ldots$ \\
\midrule

\scriptsize\texttt{[ers]}$\times$\texttt{[admin]} \newline $\to$ \textit{faculty}
& ``The university hired new''
& \textit{security guards} to guard the campus$\ldots$
& \textit{security guards} to guard the campus$\ldots$
& \textbf{\textit{faculty members}} to help improve its academic performance. \\
\midrule

\scriptsize\texttt{[ized]}$\times$\texttt{[treatment]} \newline $\to$ \textit{vaccination}
& ``Parents were urged to bring their children in for scheduled''
& visits to the hospital after a man was \textit{shot and killed} in a shooting$\ldots$
& visits to the hospital after a man was \textit{shot and killed} in a shooting$\ldots$
& \textbf{\textit{vaccinations}}, but the government has not yet taken action. \\
\midrule

\scriptsize\texttt{[secret]}$\times$\texttt{[wikileaks]} \newline $\to$ \textit{leaks}
& ``The website that published thousands of diplomatic cables is''
& now under fire for publishing a list of the \textit{names of people who leaked them}.
& now under fire for publishing a list of the \textit{names of people who leaked them}.
& now under fire for publishing a series of \textbf{\textit{leaked documents}} that show the US government$\ldots$\textbf{\textit{spy on its own citizens}}. \\

\bottomrule
\end{tabular}
\end{table*}

\begin{table*}[h]
\centering
\caption{Additional activation steering examples where vanilla SAE directions fail to shift the output from the unsteered baseline, while PolySAE directions produce a meaningful change. The first output column shows the shared continuation under no steering and vanilla SAE (both identical); the last column shows the PolySAE-steered continuation. \textbf{Bold} highlights the semantic shift introduced by PolySAE.}
\label{tab:steering_polyonly}
\small
\setlength{\tabcolsep}{4pt}
\begin{tabular}{p{2.2cm} p{2.8cm} p{5.0cm} p{5.0cm}}
\toprule
\textbf{Features $\to$ Target} & \textbf{Prompt} & \textbf{No Steering = SAE} & \textbf{PolySAE} \\
\midrule

\scriptsize\texttt{[document]}$\times$\texttt{[content]} \newline $\to$ \textit{chapters}
& ``Before reading the report, check the''
& following links: http://www.npr.org/$\ldots$
& following: The report states that the \textbf{\textit{Department of Homeland Security}} has been working with$\ldots$ \\
\midrule

\scriptsize\texttt{[Dragon]}$\times$\texttt{[steel]} \newline $\to$ \textit{Iron}
& ``The expansion pack introduced a new enemy faction called the''
& ``\textit{Mage}'' that was able to take over the world of the Elder Scrolls Online.
& ``\textbf{\textit{Grim Reaper}}'' that was able to take over the entire map. The Grim Reaper was a powerful enemy$\ldots$ \\
\midrule

\scriptsize\texttt{[earnings]}$\times$\texttt{[number]} \newline $\to$ \textit{quarterly}
& ``The stock moved sharply after the company released its''
& \textit{latest earnings report} on Tuesday. The company said it had \textit{lost} \$1.4 billion$\ldots$
& \textbf{\textit{first quarterly results}} since the financial crisis. The company said it had \textbf{\textit{sold}} more than \$1 billion$\ldots$ \\
\midrule

\scriptsize\texttt{[Star]}$\times$\texttt{[staff]} \newline $\to$ \textit{editorial}
& ``Credit for the investigative piece went to the paper's''
& editor, who wrote: ``The story was published in the NYT on Monday, and it was a great success.''
& editor, who wrote: ``The paper's \textbf{\textit{editorial staff}} has been working hard to ensure that the story is accurate.'' \\
\midrule

\scriptsize\texttt{[shooting]}$\times$\texttt{[focus]} \newline $\to$ \textit{magazine}
& ``Behind the scenes the crew was busy with the cover''
& of the new \textit{book}, which is due out in September. The book is a collection of stories$\ldots$
& of the new \textbf{\textit{issue of the magazine}}. \\
\midrule

\scriptsize\texttt{[involved]}$\times$\texttt{[support]} \newline $\to$ \textit{political}
& ``The organization's mission is to get more citizens''
& involved in the fight against \textit{climate change}.
& involved in the \textbf{\textit{political process}}. \\
\midrule

\scriptsize\texttt{[Dragon]}$\times$\texttt{[steel]} \newline $\to$ \textit{Dark Age}
& ``The fantasy novel described warriors who must''
& fight against the forces of darkness and the forces of the dark. \textit{[repetitive]}
& fight for their lives against the forces of darkness. The novel is set in the world of the \textbf{\textit{Dark Age}}$\ldots$ \\
\midrule

\scriptsize\texttt{[ers]}$\times$\texttt{[admin]} \newline $\to$ \textit{regulatory}
& ``The organization needed experienced''
& and experienced staff to help it grow. \textit{[repetitive]}
& staff to help it navigate the \textbf{\textit{complex legal and regulatory environment}}. \\

\bottomrule
\end{tabular}
\end{table*}

\section{Interaction Rank Ablation}
\label{app:rank_ablation}

\input{figures/ranks}

\Cref{fig:heatmap} examines the effect of interaction ranks on reconstruction, for fixed $R_1 = 768$ on GPT-2 Small. PolySAE achieves competitive reconstruction with modest interaction ranks ($R_2 = R_3 = 64$). Increasing ranks beyond this does not improve reconstruction, suggesting that the additional capacity is unnecessary for capturing interaction structure in this setting.

\section{Computational Overhead}
\label{app:overhead}

\begin{figure}[h]
  \centering
  \includegraphics[width=0.60\linewidth]{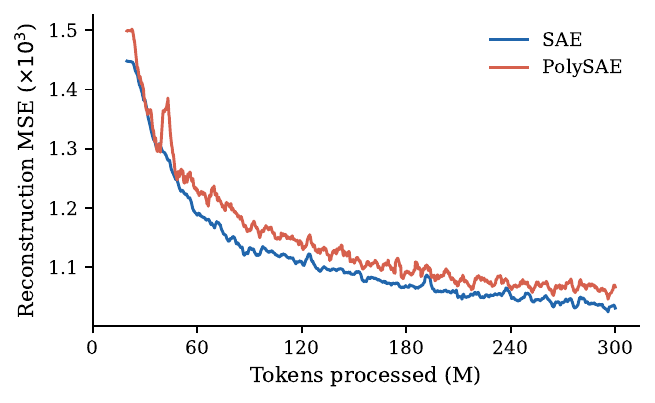}
  \caption{Reconstruction MSE over $3\times10^{8}$ training tokens.}
  \label{fig:mse_appendix}
\end{figure}

\begin{figure}[h]
  \centering
  \includegraphics[width=\linewidth]{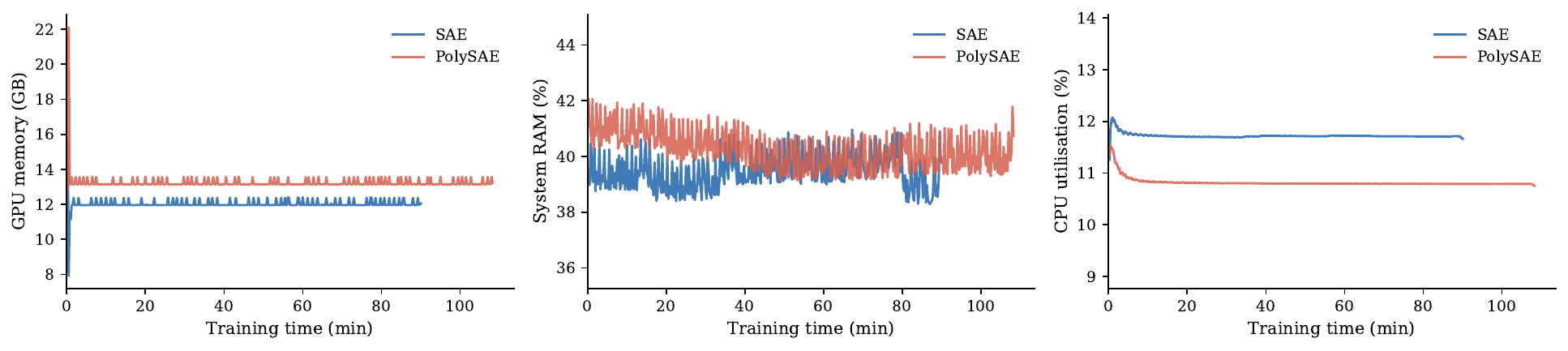}
  \caption{GPU memory, system RAM, and CPU utilization recorded during training.}
  \label{fig:system_appendix}
\end{figure}

We measure the practical cost of PolySAE's polynomial decoder against a vanilla SAE under identical training conditions on GPT-2 Small ($3\times10^{8}$ training tokens, context length 128, batch size 4096 tokens per step, single GPU). \Cref{fig:mse_appendix} shows reconstruction MSE over training, \Cref{fig:system_appendix} reports per-step system resource usage, and \Cref{tab:overhead} summarizes median resource use and total wall-clock time. PolySAE adds $+1.19$\,GB of GPU memory and $+18.2$\,min of wall-clock time, with comparable system RAM and CPU utilization. These overheads are modest relative to the gains reported in \Cref{tab:f1_scores_k1}.

\begin{table}[h]
  \centering
  \caption{Median resource usage and wall-clock time during training. $\Delta$ shows absolute difference with relative change in parentheses.}
  \label{tab:overhead}
  \footnotesize
  \setlength{\tabcolsep}{4pt}
  \begin{tabular}{lcccc}
    \toprule
    & GPU mem.\ (GB) & RAM (\%) & CPU (\%) & Wall clock (min) \\
    \midrule
    SAE     & 11.96 & 39.4 & 11.7 & 90.0  \\
    PolySAE & 13.15 & 40.2 & 10.8 & 108.2 \\
    \midrule
    $\Delta$ & $+1.19$ ($+10.0\%$) & $+0.8$ ($+2.0\%$) & $-0.9$ ($-7.7\%$) & $+18.2$ ($+20.2\%$) \\
    \bottomrule
  \end{tabular}
\end{table}

%% file: figures/ranks.tex
\begin{figure}[!t]
\centering
\begin{tikzpicture}
    \begin{groupplot}[
        group style={
            group size=2 by 1,
            horizontal sep=1.5cm,
        },
        width=0.45\textwidth,
        height=0.45\textwidth,
        view={0}{90},
        xlabel={Quadratic Rank ($R_2$)},
        ylabel={Cubic Rank ($R_3$)},
        tick label style={font=\footnotesize},
        label style={font=\small},
        title style={font=\bfseries\small},
        xtick={1,2,3,4,5},
        xticklabels={64,128,256,512,768},
        ytick={1,2,3,4,5},
        yticklabels={64,128,256,512,768},
        xmin=0.5, xmax=5.5,
        ymin=0.5, ymax=5.5,
        colorbar,
        colorbar style={width=0.12cm, tick label style={font=\tiny}},
        shader=flat,
    ]

    \nextgroupplot[
        title={Reconstruction MSE $\downarrow$},
        point meta min=0.55,
        point meta max=0.80,
    ]
    \addplot[
        matrix plot,
        mesh/rows=5,
        mesh/cols=5,
        point meta=explicit,
        nodes near coords,
        nodes near coords style={font=\tiny, anchor=center, color=black},
    ] table [meta=v] {
        x y v
        1 1 0.55
        1 2 0.60
        1 3 0.70
        1 4 0.62
        1 5 0.62

        2 1 0.56
        2 2 0.57
        2 3 0.79
        2 4 0.64
        2 5 0.61

        3 1 0.60
        3 2 0.60
        3 3 0.55
        3 4 0.60
        3 5 0.59

        4 1 0.64
        4 2 0.63
        4 3 0.63
        4 4 0.62
        4 5 0.59

        5 1 0.61
        5 2 0.62
        5 3 0.63
        5 4 0.62
        5 5 0.59
    };
    \end{groupplot}
\end{tikzpicture}
\caption{Reconstruction MSE for different $R_2$ and $R_3$ values, with $R_1=768$, using activations from GPT-2 Small.}
\label{fig:heatmap}
\end{figure}

%% file: main.bib
@article{amodei2016concrete,
  title={Concrete problems in {AI} safety},
  author={Amodei, Dario and Olah, Chris and Steinhardt, Jacob and Christiano, Paul and Schulman, John and Man{\'e}, Dan},
  journal={arXiv preprint arXiv:1606.06565},
  year={2016}
}

@article{hendrycks2021unsolved,
  title={Unsolved problems in {ML} safety},
  author={Hendrycks, Dan and Carlini, Nicholas and Schulman, John and Steinhardt, Jacob},
  journal={arXiv preprint arXiv:2109.13916},
  year={2021}
}

@article{bengio2025international,
  title={International ai safety report},
  author={Bengio, Yoshua and Mindermann, S{\"o}ren and Privitera, Daniel and Besiroglu, Tamay and Bommasani, Rishi and Casper, Stephen and Choi, Yejin and Fox, Philip and Garfinkel, Ben and Goldfarb, Danielle and others},
  journal={arXiv preprint arXiv:2501.17805},
  year={2025}
}

@inproceedings{wong2021leveraging,
  title={Leveraging sparse linear layers for debuggable deep networks},
  author={Wong, Eric and Santurkar, Shibani and Madry, Aleksander},
  booktitle={International Conference on Machine Learning},
  pages={11205--11216},
  year={2021},
  organization={PMLR}
}

@inproceedings{rimsky2024steering,
  title={Steering llama 2 via contrastive activation addition},
  author={Rimsky, Nina and Gabrieli, Nick and Schulz, Julian and Tong, Meg and Hubinger, Evan and Turner, Alexander},
  booktitle={Proceedings of the 62nd Annual Meeting of the Association for Computational Linguistics (Volume 1: Long Papers)},
  pages={15504--15522},
  year={2024}
}

@misc{lindsey2025biology,
  author = {Lindsey, Jack and Gurnee, Wes and Ameisen, Emmanuel and Chen, Brian and Pearce, Adam and Turner, Nicholas L. and Citro, Craig and Abrahams, David and Carter, Shan and Hosmer, Basil and Marcus, Jonathan and Sklar, Michael and Templeton, Adly and Bricken, Trenton and McDougall, Callum and Cunningham, Hoagy and Henighan, Thomas and Jermyn, Adam and Jones, Andy and Persic, Andrew and Qi, Zhenyi and Thompson, T. Ben and Zimmerman, Sam and Rivoire, Kelley and Conerly, Thomas and Olah, Chris and Batson, Joshua},
  title  = {On the Biology of a Large Language Model},
  year   = {2025},
  howpublished = {Transformer Circuits Thread},
  url    = {https://transformer-circuits.pub/2025/attribution-graphs/biology.html}
}

@article{bereska2024mechanistic,
title={Mechanistic Interpretability for {AI} Safety - A Review},
author={Leonard Bereska and Stratis Gavves},
journal={Transactions on Machine Learning Research},
issn={2835-8856},
year={2024},
url={https://openreview.net/forum?id=ePUVetPKu6},
note={Survey Certification, Expert Certification}
}

@article{mallat1993matching,
  title={Matching pursuits with time-frequency dictionaries},
  author={Mallat, St{\'e}phane G and Zhang, Zhifeng},
  journal={IEEE Transactions on signal processing},
  volume={41},
  number={12},
  pages={3397--3415},
  year={1993},
  publisher={IEEE}
}

@article{hinton2006reducing,
  title={Reducing the dimensionality of data with neural networks},
  author={Hinton, Geoffrey E and Salakhutdinov, Ruslan R},
  journal={Science},
  volume={313},
  number={5786},
  pages={504--507},
  year={2006},
  publisher={American Association for the Advancement of Science}
}

@article{lee2007sparse,
  title={Sparse deep belief net model for visual area V2},
  author={Lee, Honglak and Ekanadham, Chaitanya and Ng, Andrew},
  journal={Advances in Neural Information Processing Systems},
  volume={20},
  year={2007}
}

@article{konda2014zero,
  title={Zero-bias autoencoders and the benefits of co-adapting features},
  author={Konda, Kishore and Memisevic, Roland and Krueger, David},
  journal={arXiv preprint arXiv:1402.3337},
  year={2014}
}

@article{olshausen1997sparse,
  title={Sparse coding with an overcomplete basis set: A strategy employed by {V1}?},
  author={Olshausen, Bruno A and Field, David J},
  journal={Vision Research},
  volume={37},
  number={23},
  pages={3311--3325},
  year={1997},
  publisher={Elsevier}
}

@inproceedings{arora2015simple,
  title={Simple, efficient, and neural algorithms for sparse coding},
  author={Arora, Sanjeev and Ge, Rong and Ma, Tengyu and Moitra, Ankur},
  booktitle={Conference on Learning Theory (COLT)},
  pages={113--149},
  year={2015},
  organization={PMLR}
}

@article{bao2016dictionary,
  title={Dictionary learning for sparse coding: Algorithms and analysis},
  author={Bao, Chenglong and Ji, Hui and Quan, Yuhui and Shen, Zuowei},
  journal={IEEE Transactions on Pattern Analysis and Machine Intelligence},
  volume={38},
  number={7},
  pages={1356--1369},
  year={2016},
  publisher={IEEE}
}

@inproceedings{
engelsnot,
title={Not All Language Model Features Are One-Dimensionally Linear},
author={Joshua Engels and Eric J Michaud and Isaac Liao and Wes Gurnee and Max Tegmark},
booktitle={The Thirteenth International Conference on Learning Representations},
year={2025},
url={https://openreview.net/forum?id=d63a4AM4hb}
}

@article{elhage2022toy,
  title={Toy models of superposition},
  author={Elhage, Nelson and Hume, Tristan and Olsson, Catherine and Schiefer, Nicholas and Henighan, Tom and Kravec, Shauna and Hatfield-Dodds, Zac and Lasenby, Robert and Drain, Dawn and Chen, Carol and others},
  journal={Transformer Circuits Thread},
  year={2022},
  note={https://transformer-circuits.pub/2022/toy\_model/index.html}
}

@misc{bricken2023monosemanticity,
  author = {Bricken, Trenton and Templeton, Adly and Batson, Joshua and Chen, Brian and Jermyn, Adam and Conerly, Tom and Turner, Nicholas L. and Anil, Cem and Denison, Carson and Askell, Amanda and Lasenby, Robert and Wu, Yifan and Kravec, Shauna and Schiefer, Nicholas and Maxwell, Tim and Joseph, Nicholas and Hatfield-Dodds, Zac and Tamkin, Alex and Nguyen, Karina and McLean, Brayden and Burke, Josiah E. and Hume, Tristan and Carter, Shan and Henighan, Tom and Olah, Christopher},
  title  = {Towards Monosemanticity: Decomposing Language Models with Dictionary Learning},
  year   = {2023},
  howpublished = {Transformer Circuits Thread},
  url    = {https://transformer-circuits.pub/2023/monosemantic-features/index.html}
}

@inproceedings{
huben2024sparse,
title={Sparse Autoencoders Find Highly Interpretable Features in Language Models},
author={Robert Huben and Hoagy Cunningham and Logan Riggs Smith and Aidan Ewart and Lee Sharkey},
booktitle={The Twelfth International Conference on Learning Representations},
year={2024},
url={https://openreview.net/forum?id=F76bwRSLeK}
}

@inproceedings{gao2025scaling, title={Scaling and Evaluating Sparse Autoencoders}, author={Gao, Leo and Dupre la Tour, Tom and Tillman, Henk and Goh, Gabriel and Troll, Rajan and Radford, Alec and Sutskever, Ilya and Leike, Jan and Wu, Jeffrey}, booktitle={The Thirteenth International Conference on Learning Representations (ICLR)}, year={2025}, url={https://openreview.net/forum?id=tcsZt9ZNKD} }

@inproceedings{bussmann2024batchtopk,
  title={{BatchTopK} Sparse Autoencoders},
  author={Bussmann, Bart and Leask, Patrick and Nanda, Neel},
  booktitle={NeurIPS Workshop on Scientific Methods for Understanding Deep Learning},
  year={2024},
  url={https://openreview.net/forum?id=d4dpOCqybL}
}

@inproceedings{bussmann2025matryoshka,
  title={Learning Multi-Level Features with {Matryoshka} Sparse Autoencoders},
  author={Bussmann, Bart and Nabeshima, Noa and Karvonen, Adam and Nanda, Neel},
  booktitle={Proceedings of the 42nd International Conference on Machine Learning (ICML)},
  series={Proceedings of Machine Learning Research},
  volume={267},
  year={2025},
  publisher={PMLR}
}

@article{rajamanoharan2024improving,
  title={Improving dictionary learning with gated sparse autoencoders},
  author={Rajamanoharan, Senthooran and Conmy, Arthur and Smith, Lewis and Lieberum, Tom and Varma, Vikrant and Kram{\'a}r, J{\'a}nos and Shah, Rohin and Nanda, Neel},
  journal={arXiv preprint arXiv:2404.16014},
  year={2024}
}

@article{rajamanoharan2024jumping,
  title={Jumping ahead: Improving reconstruction fidelity with jumprelu sparse autoencoders},
  author={Rajamanoharan, Senthooran and Lieberum, Tom and Sonnerat, Nicolas and Conmy, Arthur and Varma, Vikrant and Kram{\'a}r, J{\'a}nos and Nanda, Neel},
  journal={arXiv preprint arXiv:2407.14435},
  year={2024}
}

@misc{templeton2024scaling,
  author = {Templeton, Adly and Conerly, Tom and Marcus, Jonathan and Lindsey, Jack and Bricken, Trenton and Chen, Brian and Pearce, Adam and Citro, Craig and Ameisen, Emmanuel and Jones, Andy and Cunningham, Hoagy and Turner, Nicholas L. and McDougall, Callum and MacDiarmid, Monte and Freeman, C. Daniel and Sumers, Theodore R. and Rees, Edward and Batson, Joshua and Jermyn, Adam and Carter, Shan and Olah, Christopher and Henighan, Tom},
  title  = {Scaling Monosemanticity: Extracting Interpretable Features from {Claude} 3 {Sonnet}},
  year   = {2024},
  howpublished = {Transformer Circuits Thread},
  url    = {https://transformer-circuits.pub/2024/scaling-monosemanticity/index.html}
}

@book{haspelmath2013understanding,
  title={Understanding morphology},
  author={Haspelmath, Martin and Sims, Andrea},
  year={2013},
  publisher={Routledge}
}

@inproceedings{csordas2024recurrent,
  title={Recurrent Neural Networks Learn to Store and Generate Sequences using Non-Linear Representations},
  author={Csord{\'a}s, R{\'o}bert and Potts, Christopher and Manning, Christopher D and Geiger, Atticus},
  booktitle={Proceedings of the 7th BlackboxNLP Workshop: Analyzing and Interpreting Neural Networks for NLP},
  pages={248--262},
  year={2024}
}

@article{alain2016understanding,
  title={Understanding intermediate layers using linear classifier probes},
  author={Alain, Guillaume and Bengio, Yoshua},
  journal={arXiv preprint arXiv:1610.01644},
  year={2016}
}

@article{belinkov2022probing,
  title={Probing classifiers: Promises, shortcomings, and advances},
  author={Belinkov, Yonatan},
  journal={Computational Linguistics},
  volume={48},
  number={1},
  pages={207--219},
  year={2022},
  publisher={MIT Press}
}

@inproceedings{meng2022locating,
  title={Locating and editing factual associations in {GPT}},
  author={Meng, Kevin and Bau, David and Andonian, Alex and Belinkov, Yonatan},
  booktitle={Advances in Neural Information Processing Systems (NeurIPS)},
  volume={35},
  pages={17359--17372},
  year={2022}
}

@inproceedings{koehn-2005-europarl,
    title = "{E}uroparl: A Parallel Corpus for Statistical Machine Translation",
    author = "Koehn, Philipp",
    booktitle = "Proceedings of Machine Translation Summit X: Papers",
    month = sep # " 13-15",
    year = "2005",
    address = "Phuket, Thailand",
    pages = "79--86",
}

@article{hou2024bridging,
  title={Bridging Language and Items for Retrieval and Recommendation},
  author={Hou, Yupeng and Li, Jiacheng and He, Zhankui and Yan, An and Chen, Xiusi and McAuley, Julian},
  journal={arXiv preprint arXiv:2403.03952},
  year={2024}
}

@misc{codeparrot_github_code,
  title        = {GitHub Code Dataset},
  author       = {{CodeParrot}},
  year         = {2022},
  howpublished = {\url{https://huggingface.co/datasets/codeparrot/github-code}},
}

@book{volterra1959theory,
  title={Theory of Functionals and of Integral and Integro-Differential Equations},
  author={Volterra, Vito},
  year={1959},
  publisher={Dover Publications},
  note={Originally published 1930}
}

@article{chrysos2021deep,
  author  = {Chrysos, Grigorios G. and Moschoglou, Stylianos and Bouritsas, Giorgos and Deng, Jiankang and Panagakis, Yannis and Zafeiriou, Stefanos},
  title   = {Deep Polynomial Neural Networks},
  journal = {IEEE Transactions on Pattern Analysis and Machine Intelligence},
  volume  = {44},
  number  = {8},
  pages   = {4021--4034},
  year    = {2022},
  doi     = {10.1109/TPAMI.2021.3058891}
}

@inproceedings{shin1991pi,
  title={The pi-sigma network: An efficient higher-order neural network for pattern classification and function approximation},
  author={Shin, Yoan and Ghosh, Joydeep},
  booktitle={IJCNN-91-Seattle international joint conference on neural networks},
  volume={1},
  pages={13--18},
  year={1991},
  organization={IEEE}
}

@inproceedings{chrysos2020picvpr,
  title={P-nets: Deep polynomial neural networks},
  author={Chrysos, Grigorios G and Moschoglou, Stylianos and Bouritsas, Giorgos and Panagakis, Yannis and Deng, Jiankang and Zafeiriou, Stefanos},
  booktitle={Proceedings of the IEEE/CVF Conference on Computer Vision and Pattern Recognition},
  pages={7325--7335},
  year={2020}
}

@article{chrysos2021conditional,
  title={Conditional generation using polynomial expansions},
  author={Chrysos, Grigorios and Georgopoulos, Markos and Panagakis, Yannis},
  journal={Advances in Neural Information Processing Systems},
  volume={34},
  pages={28390--28404},
  year={2021}
}

@article{fodor1988connectionism,
  title={Connectionism and cognitive architecture: A critical analysis},
  author={Fodor, Jerry A and Pylyshyn, Zenon W},
  journal={Cognition},
  volume={28},
  number={1-2},
  pages={3--71},
  year={1988},
  publisher={Elsevier}
}

@article{smolensky1990tensor,
  title={Tensor product variable binding and the representation of symbolic structures in connectionist systems},
  author={Smolensky, Paul},
  journal={Artificial Intelligence},
  volume={46},
  number={1-2},
  pages={159--216},
  year={1990},
  publisher={Elsevier}
}

@inproceedings{carreira2012semantic,
  title={Semantic segmentation with second-order pooling},
  author={Carreira, Jo{\~a}o and Caseiro, Rui and Batista, Jorge and Sminchisescu, Cristian},
  booktitle={European Conference on Computer Vision (ECCV)},
  pages={430--443},
  year={2012},
  publisher={Springer}
}

@inproceedings{gao2016compact,
  title={Compact bilinear pooling},
  author={Gao, Yang and Beijbom, Oscar and Zhang, Ning and Darrell, Trevor},
  booktitle={IEEE Conference on Computer Vision and Pattern Recognition (CVPR)},
  pages={317--326},
  year={2016}
}

@article{grasedyck2013literature,
  title={A literature survey of low-rank tensor approximation techniques},
  author={Grasedyck, Lars and Kressner, Daniel and Tobler, Christine},
  journal={GAMM-Mitteilungen},
  volume={36},
  number={1},
  pages={53--78},
  year={2013},
  publisher={Wiley}
}

@book{absil2008optimization,
  author    = {Absil, P.-A. and Mahony, R. and Sepulchre, R.},
  title     = {Optimization Algorithms on Matrix Manifolds},
  publisher = {Princeton University Press},
  address   = {Princeton, NJ},
  year      = {2008},
  pages     = {xvi+224},
  isbn      = {978-0-691-13298-3}
}

@article{edelman1998geometry,
  title={The geometry of algorithms with orthogonality constraints},
  author={Edelman, Alan and Arias, Tom{\'a}s A and Smith, Steven T},
  journal={SIAM Journal on Matrix Analysis and Applications},
  volume={20},
  number={2},
  pages={303--353},
  year={1998},
  publisher={SIAM}
}

@article{bonnabel2013stochastic,
  title={Stochastic gradient descent on {R}iemannian manifolds},
  author={Bonnabel, Silv{\`e}re},
  journal={IEEE Transactions on Automatic Control},
  volume={58},
  number={9},
  pages={2217--2229},
  year={2013},
  doi={10.1109/TAC.2013.2254619},
  publisher={IEEE}
}

@article{hyvarinen2000independent,
  title={Independent component analysis: Algorithms and applications},
  author={Hyv{\"a}rinen, Aapo and Oja, Erkki},
  journal={Neural Networks},
  volume={13},
  number={4-5},
  pages={411--430},
  year={2000},
  publisher={Elsevier}
}

@book{mason2002chebyshev,
  author    = {Mason, J. C. and Handscomb, D. C.},
  title     = {Chebyshev Polynomials},
  publisher = {Chapman and Hall/CRC},
  address   = {Boca Raton, FL},
  year      = {2002},
  isbn      = {978-0-849-30355-5}
}

@inproceedings{rendle2010factorization,
  title={Factorization Machines},
  author={Rendle, Steffen},
  booktitle={2010 IEEE International Conference on Data Mining},
  pages={995--1000},
  year={2010},
  organization={IEEE}
}

@inproceedings{blondel2016higher,
  title={Higher-Order Factorization Machines},
  author={Blondel, Mathieu and Fujino, Akinori and Ueda, Naonori and Ishihata, Masakazu},
  booktitle={Advances in Neural Information Processing Systems},
  volume={29},
  pages={3351--3359},
  year={2016}
}

@inproceedings{kim2017hadamard,
  title={Hadamard Product for Low-Rank Bilinear Pooling},
  author={Kim, Jin-Hwa and On, Kyoung-Woon and Lim, Woosang and Kim, Jeonghee and Ha, Jung-Woo and Zhang, Byoung-Tak},
  booktitle={International Conference on Learning Representations},
  year={2017}
}

@incollection{partee1995lexical,
  title={Lexical Semantics and Compositionality},
  author={Partee, Barbara H.},
  booktitle={An Invitation to Cognitive Science: Language},
  editor={Gleitman, Lila R. and Liberman, Mark},
  volume={1},
  pages={311--360},
  year={1995},
  publisher={MIT Press},
  address={Cambridge, MA}
}

@InProceedings{karvonen2025saebench,
  title     = {{SAEBench}: A Comprehensive Benchmark for Sparse Autoencoders in Language Model Interpretability},
  author    = {Karvonen, Adam and Rager, Can and Lin, Johnny and Tigges, Curt and Bloom, Joseph Isaac and Chanin, David and Lau, Yeu-Tong and Farrell, Eoin and McDougall, Callum Stuart and Ayonrinde, Kola and Till, Demian and Wearden, Matthew and Conmy, Arthur and Marks, Samuel and Nanda, Neel},
  booktitle = {Proceedings of the 42nd International Conference on Machine Learning},
  series    = {Proceedings of Machine Learning Research},
  volume    = {267},
  pages     = {29223--29264},
  year      = {2025},
  publisher = {PMLR},
  url       = {https://proceedings.mlr.press/v267/karvonen25a.html}
}

@InProceedings{biderman2023pythia,
  title     = {{P}ythia: A Suite for Analyzing Large Language Models Across Training and Scaling},
  author    = {Biderman, Stella and Schoelkopf, Hailey and Anthony, Quentin Gregory and Bradley, Herbie and O'Brien, Kyle and Hallahan, Eric and Khan, Mohammad Aflah and Purohit, Shivanshu and Prashanth, Usvsn Sai and Raff, Edward and Skowron, Aviya and Sutawika, Lintang and Van Der Wal, Oskar},
  booktitle = {Proceedings of the 40th International Conference on Machine Learning},
  series    = {Proceedings of Machine Learning Research},
  volume    = {202},
  pages     = {2397--2430},
  year      = {2023},
  publisher = {PMLR},
  url       = {https://proceedings.mlr.press/v202/biderman23a.html}
}

@techreport{radford2019gpt2,
  title       = {Language Models are Unsupervised Multitask Learners},
  author      = {Radford, Alec and Wu, Jeffrey and Child, Rewon and Luan, David and Amodei, Dario and Sutskever, Ilya},
  year        = {2019},
  institution = {OpenAI},
  url         = {https://cdn.openai.com/better-language-models/language_models_are_unsupervised_multitask_learners.pdf}
}

@techreport{gemma2report,
  title       = {{G}emma 2: Improving Open Language Models at a Practical Size},
  author      = {{Gemma Team}},
  year        = {2024},
  institution = {Google DeepMind},
  url         = {https://storage.googleapis.com/deepmind-media/gemma/gemma-2-report.pdf}
}

@article{gao2021pile,
  title   = {The {P}ile: An 800{GB} Dataset of Diverse Text for Language Modeling},
  author  = {Gao, Leo and Biderman, Stella and Black, Sid and Golding, Laurence and Hoppe, Travis and Foster, Charles and Phang, Jason and He, Horace and Thite, Anish and Nabeshima, Noa and Presser, Shawn and Leahy, Connor},
  journal = {arXiv preprint arXiv:2101.00027},
  year    = {2021}
}

@inproceedings{
dunefsky2024transcoders,
title={Transcoders find interpretable {LLM} feature circuits},
author={Jacob Dunefsky and Philippe Chlenski and Neel Nanda},
booktitle={The Thirty-eighth Annual Conference on Neural Information Processing Systems},
year={2024},
url={https://openreview.net/forum?id=J6zHcScAo0}
}

@misc{gokaslan2019openwebtext,
  title        = {{O}pen{W}eb{T}ext Corpus},
  author       = {Gokaslan, Aaron and Cohen, Vanya and Pavlick, Ellie and Tellex, Stefanie},
  year         = {2019},
  howpublished = {Zenodo},
  url          = {https://zenodo.org/records/3834942}
}

@inproceedings{dearteaga2019biasinbios,
  title     = {Bias in Bios: A Case Study of Semantic Representation Bias in a High-Stakes Setting},
  author    = {De-Arteaga, Maria and Romanov, Alexey and Wallach, Hanna and Chayes, Jennifer and Borgs, Christian and Chouldechova, Alexandra and Geyik, Sahin Cem and Kenthapadi, Krishnaram and Kalai, Adam Tauman},
  booktitle = {Proceedings of the Conference on Fairness, Accountability, and Transparency ({FAT}* '19)},
  pages     = {120--128},
  year      = {2019},
  publisher = {ACM},
  doi       = {10.1145/3287560.3287572}
}

@inproceedings{zhang2015charcnn,
  title     = {Character-level Convolutional Networks for Text Classification},
  author    = {Zhang, Xiang and Zhao, Junbo and LeCun, Yann},
  booktitle = {Advances in Neural Information Processing Systems},
  volume    = {28},
  year      = {2015},
  url       = {https://proceedings.neurips.cc/paper/2015/file/250cf8b51c773f3f8dc8b4be867a9a02-Paper.pdf}
}

@misc{bloom2024saetrainingcodebase,
  title        = {{SAELens}},
  author       = {Bloom, Joseph and Tigges, Curt and Duong, Anthony and Chanin, David},
  year         = {2024},
  howpublished = {\url{https://github.com/jbloomAus/SAELens}},
  note         = {GitHub repository}
}

@inproceedings{jayakumar2020multiplicative,
  title={Multiplicative interactions and where to find them},
  author={Jayakumar, Siddhant M and Czarnecki, Wojciech M and Menick, Jacob and Schwarz, Jonathan and Rae, Jack and Osindero, Simon and Teh, Yee Whye and Harley, Tim and Pascanu, Razvan},
  booktitle={International Conference on Learning Representations},
  year={2020}
}

@article{chrysos2025hadamard,
  title={Hadamard product in deep learning: Introduction, Advances and Challenges},
  author={Chrysos, Grigorios G and Wu, Yongtao and Pascanu, Razvan and Torr, Philip and Cevher, Volkan},
  journal={IEEE Transactions on Pattern Analysis and Machine Intelligence},
  year={2025},
  publisher={IEEE}
}

@inproceedings{freeman1997learning,
  title={Learning bilinear models for two-factor problems in vision},
  author={Freeman, William T and Tenenbaum, Joshua B},
  booktitle={Proceedings of IEEE Computer Society Conference on Computer Vision and Pattern Recognition},
  pages={554--560},
  year={1997},
  organization={IEEE}
}

@article{tenenbaum1996separating,
  title={Separating style and content},
  author={Tenenbaum, Joshua and Freeman, William},
  journal={Advances in Neural Information Processing Systems},
  volume={9},
  year={1996}
}

@inproceedings{perez2018film,
  title={Film: Visual reasoning with a general conditioning layer},
  author={Perez, Ethan and Strub, Florian and De Vries, Harm and Dumoulin, Vincent and Courville, Aaron},
  booktitle={Proceedings of the AAAI conference on artificial intelligence},
  volume={32},
  number={1},
  year={2018}
}

@inproceedings{dumoulin2017a,
title={A Learned Representation For Artistic Style},
author={Vincent Dumoulin and Jonathon Shlens and Manjunath Kudlur},
booktitle={International Conference on Learning Representations},
year={2017},
url={https://openreview.net/forum?id=BJO-BuT1g}
}

@misc{shazeer2020glu,
      title={{GLU} Variants Improve Transformer}, 
      author={Noam Shazeer},
      year={2020},
      eprint={2002.05202},
      archivePrefix={arXiv},
      primaryClass={cs.LG},
      url={https://arxiv.org/abs/2002.05202}, 
}

@inproceedings{
oldfield2025towards,
title={Towards Interpretability Without Sacrifice: Faithful Dense Layer Decomposition with Mixture of Decoders},
author={James Oldfield and Shawn Im and Sharon Li and Mihalis Nicolaou and Ioannis Patras and Grigorios Chrysos},
booktitle={The Thirty-ninth Annual Conference on Neural Information Processing Systems},
year={2025},
url={https://openreview.net/forum?id=jcvX8XFNqX}
}

@inproceedings{pearce2025bilinear,
title={Bilinear {MLP}s enable weight-based mechanistic interpretability},
author={Michael T Pearce and Thomas Dooms and Alice Rigg and Jose Oramas and Lee Sharkey},
booktitle={The Thirteenth International Conference on Learning Representations},
year={2025},
url={https://openreview.net/forum?id=gI0kPklUKS}
}

@inproceedings{gauderis2025compositionality,
  title={Compositionality Unlocks Deep Interpretable Models},
  author={Gauderis, Ward and Dooms, Thomas},
  booktitle={Connecting Low-Rank Representations in AI: At the 39th Annual AAAI Conference on Artificial Intelligence},
  year={2025}
}

@inproceedings{dooms2025bilinearauto,
title={Finding Manifolds With Bilinear Autoencoders},
author={Thomas Dooms and Ward Gauderis},
booktitle={Mechanistic Interpretability Workshop at NeurIPS 2025},
year={2025},
url={https://openreview.net/forum?id=ybJXIh4vcF}
}

@inproceedings{babiloni2021poly,
  title={Poly-nl: Linear complexity non-local layers with 3rd order polynomials},
  author={Babiloni, Francesca and Marras, Ioannis and Kokkinos, Filippos and Deng, Jiankang and Chrysos, Grigorios and Zafeiriou, Stefanos},
  booktitle={Proceedings of the IEEE/CVF international conference on computer vision},
  pages={10518--10528},
  year={2021}
}

@inproceedings{chrysos2022augmenting,
  title={Augmenting deep classifiers with polynomial neural networks},
  author={Chrysos, Grigorios G and Georgopoulos, Markos and Deng, Jiankang and Kossaifi, Jean and Panagakis, Yannis and Anandkumar, Anima},
  booktitle={European Conference on Computer Vision},
  pages={692--716},
  year={2022},
  organization={Springer}
}

@inproceedings{zhang2019encrypted,
  title={Encrypted speech recognition using deep polynomial networks},
  author={Zhang, Shi-Xiong and Gong, Yifan and Yu, Dong},
  booktitle={ICASSP 2019-2019 IEEE International Conference on Acoustics, Speech and Signal Processing (ICASSP)},
  pages={5691--5695},
  year={2019},
  organization={IEEE}
}

@article{dubey2022scalable,
  title={Scalable interpretability via polynomials},
  author={Dubey, Abhimanyu and Radenovic, Filip and Mahajan, Dhruv},
  journal={Advances in neural information processing systems},
  volume={35},
  pages={36748--36761},
  year={2022}
}

@inproceedings{oldfield2026beyond,
  title={Beyond Linear Probes: Dynamic Safety Monitoring for Language Models},
  author={Oldfield, James and Torr, Philip and Patras, Ioannis and Bibi, Adel and Barez, Fazl},
  booktitle={The Fourteenth International Conference on Learning Representations},
  year={2026}
}

@book{asher2011lexical,
  author    = {Nicholas Asher},
  title     = {Lexical Meaning in Context: A Web of Words},
  publisher = {Cambridge University Press},
  year      = {2011}
}

@misc{parsons1990events,
  title={Events in the Semantics of English},
  author={Parsons, Terence},
  year={1990},
  publisher={Cambridge, Ma: MIT Press}
}

@inproceedings{marks2024geometry,
  title     = {The Geometry of Truth: Emergent Linear Structure in Large Language Model Representations of True/False Datasets},
  author    = {Marks, Samuel and Tegmark, Max},
  booktitle = {Conference on Language Modeling (COLM)},
  year      = {2024}
}

@inproceedings{wu2025axbench,
  title     = {{AxBench}: Steering {LLM}s? {E}ven Simple Baselines Outperform Sparse Autoencoders},
  author    = {Wu, Zhengxuan and Arora, Aryaman and Geiger, Atticus and Wang, Zheng and Huang, Jing and Jurafsky, Dan and Manning, Christopher D. and Potts, Christopher},
  booktitle = {Proceedings of the 42nd International Conference on Machine Learning (ICML)},
  series    = {Proceedings of Machine Learning Research},
  publisher = {PMLR},
  year      = {2025},
  note      = {Spotlight}
}
